\pdfoutput=1

\documentclass[11pt]{article}

\usepackage[final]{acl}

\usepackage{times}
\usepackage{latexsym}

\usepackage[T1]{fontenc}

\usepackage[utf8]{inputenc}

\usepackage{microtype}

\usepackage{inconsolata}

\usepackage{graphicx}

\usepackage{booktabs}
\usepackage{multirow}
\usepackage{multicol}
\usepackage{amsmath,amsfonts,amsthm,amssymb,amsopn,bm}
\usepackage{xspace}
\usepackage{tcolorbox}
\usepackage[ruled,linesnumbered]{algorithm2e}

\usepackage{subcaption}

\newcommand{\F}{Fig.}
\renewcommand{\F}{Figure}
\newcommand{\T}{Table}
\renewcommand{\S}{Section}
\newcommand{\A}{Algorithm}
\newcommand{\AP}{Appendix}

\newcommand{\tool}{\textsc{Portia}\xspace}
\newcommand{\name}{\textsc{Portia}\xspace}
\newcommand{\parh}[1]{\noindent\textbf{#1}}

\title{Split and Merge: Aligning Position Biases in Large Language Model based Evaluators}
\title{Split and Merge: Aligning Position Biases in LLM-based Evaluators}

\author{Zongjie Li$^1$, Chaozheng Wang$^2$, Pingchuan Ma$^1$, \\
{\bf Daoyuan Wu}$^{1\dagger}$, \bf{Shuai Wang}$^{1\dagger}$, \bf{Cuiyun Gao}$^2$, \bf{Yang Liu}$^3$ \\
  $^1$Hong Kong University of Science and Technology\\ $^2$Harbin Institute of Technology, $^3$Nanyang Technological University \\
  \texttt{\{zligo,pmaab,daoyuan,shuaiw\}@cse.ust.hk, \{yangliu\}@ntu.edu.sg,} \\
  \texttt{wangchaozheng@stu.hit.edu.cn, gaocuiyun@hit.edu.cn}
}

\begin{document}
\maketitle
\begin{abstract}
  Large language models (LLMs) have shown promise as automated evaluators for
  assessing the quality of answers generated by AI systems. However, LLM-based evaluators exhibit
  position bias, or inconsistency, when used to evaluate candidate answers in
  pairwise comparisons, favoring either the first or second answer regardless of content.
  To address this limitation, we propose \tool, an alignment-based system
  designed to mimic human comparison strategies to calibrate position bias 
  in a lightweight yet effective manner.
  Specifically, \tool splits the answers into multiple segments, taking into account both length and semantics, 
  and merges them back into a single prompt for evaluation by LLMs.
  Extensive experiments with six LLMs on 11,520 answer pairs demonstrate that \tool\ markedly enhances the consistency rates for all models and forms of comparison tested, achieving an average relative improvement of 47.46\%. 
  It also enables \tool-enhanced GPT-3.5 to achieve agreement rates with humans comparable to GPT-4 and elevates GPT-4's consistency rate up to 98\%.
  Subsequent human evaluations indicate that the \name-enhanced GPT-3.5 model can even surpass standalone GPT-4 in terms of alignment with human evaluators, highlighting \name's ability to correct position bias, improve LLM consistency, and boost performance while keeping cost efficiency.
\end{abstract}

\renewcommand{\thefootnote}{\fnsymbol{footnote}}
\footnotetext[2]{Corresponding authors.}  
\renewcommand{\thefootnote}{\arabic{footnote}}

\section{Introduction}
\label{sec:intro}

Recent advances in large language models (LLMs) have achieved remarkable results on various tasks, sometimes even exceeding human performance~\citep{kojima2022large,thapa2023humans}.
However, assessing the quality of LLM-generated answers poses challenges.
Specifically, n-gram matching metrics like BLEU~\citep{papineni2002bleu} can quantify token-level overlap with reference texts but fall short in evaluating semantic quality.
While human evaluators provide more accurate and valuable feedback, often considered the ``gold standards,'' their scalability is generally low, given that they are costly and time-consuming.
As a result, there emerges a growing need for automated evaluation methods that reliably align with human yet remain efficient and cost-effective.

Recently, researchers have investigated the use of powerful LLMs like GPT-4~\citep{openai2023gpt4} to evaluate the quality of text generated in response to open-ended questions~\citep{zheng2023judging}.
Notably, robust LLM evaluators such as GPT-4 have been shown to align remarkably well with both controlled and crowdsourced human preferences, achieving over 60\% agreement~\citep{wang2023large}.
These studies suggest that LLMs can emulate human evaluations, offering a scalable and transparent alternative to the expensive and time-intensive human assessment of text quality.

\begin{figure*}[!htpb]
    \centering
    \includegraphics[width=1.0\textwidth]{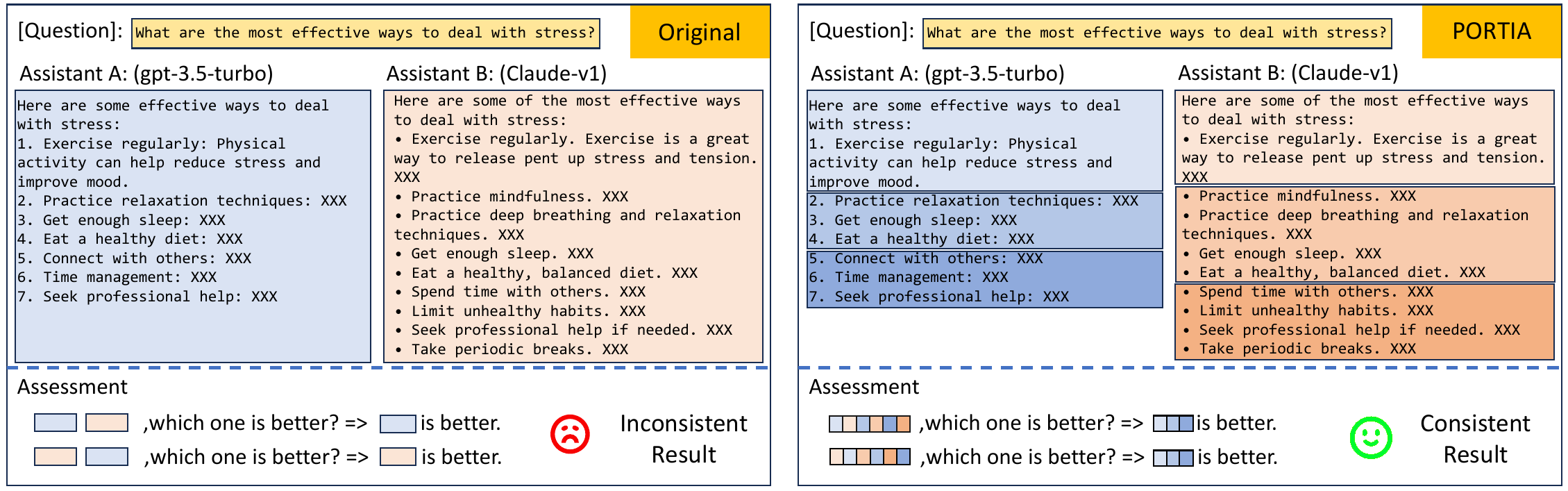}
    \caption{A sample pairwise LLM-based evaluation improved by \tool. \textit{Left:} The original evaluation exhibiting inconsistency. \textit{Right:} Consistent evaluation after applying \tool.
    Details of the answers, comparison forms, and evaluation contents have been simplified or omitted for clarity. 
    An example of the detailed prompt is given in~\T~\ref{tab:tool-template}.
    It is worth noting that the answers from different assistants may differ significantly from each other.
    }
    \label{fig:example}
\end{figure*}

While LLMs have advanced capabilities, they are not flawless evaluators and have been identified to possess certain biases.
One notable bias is the position bias~\citep{zheng2023judging,wang2023large}, in which an LLM might prefer either the first or second answer in a pairwise comparison, regardless of its content, as illustrated in~\F~\ref{fig:example}.
Even the state-of-the-art GPT-4 model is not immune to position bias~\citep{zheng2023judging,wang2023large,zhang2023wider,zeng2023evaluating}, and the behavior of its various versions can be inconsistent over time~\citep{chen2023chatgpt}. 
Moreover, owing to pronounced position biases in less-powerful GPT models, much of the prior research~\citep{zheng2023judging,zhang2023wider} has been compelled to use the expensive GPT-4 for LLM evaluations, emphasizing the necessity for a more cost-effective approach to large-scale assessments.

To address these limitations, we propose \tool\footnote{The reason for the naming is provided in \AP~\ref{app:name-reason}.}, an alignment-based system designed to calibrate position bias.
Inspired by human long-text reading strategies~\citep{ratnasari2023students}, \tool\ splits the answers into multiple segments, aligns similar content across candidate answers, and then merges them back into a single prompt to feed to LLM evaluators.
Specifically, \name first identifies possible split positions at sentence boundaries within each answer.
It then conducts a length alignment between the candidates to generate segments of roughly equal length across answers.
If this length alignment does not yield a consistent verdict, \tool further undertakes an iterative semantic alignment to identify the optimal split positions, enabling the merging of segments across candidates.
Since this lightweight approach does not require changes to the models themselves, \name is readily adaptable to enhance a variety of LLM evaluators for improved evaluation consistency.

We conducted comprehensive experiments using six LLMs as evaluators to assess 11,520 answer pairs across three prevalent pairwise comparison forms.
Our results show that \tool markedly boosts consistency rates for all the tested models and templates, achieving an average relative improvement of 47.46\% and rectifying an average of 62.31\% of the initially inconsistent cases.
Furthermore, \name addresses between 36\% and 86\% (over 80\% for two-thirds of the comparison templates) of the position bias occurrences within the GPT-4 model, elevating its consistency rate up to 98\%.
Moreover, efficiency and cost evaluations indicate that \name enables the less advanced GPT-3.5 model to achieve 88\% agreement with the state-of-the-art GPT-4 model at merely 9.57\% of the cost.
Additionally, a user study involving five human participants demonstrated enhanced agreement between \tool-optimized evaluators and human evaluators.
Remarkably, the agreement of human evaluators with \tool-enhanced GPT-3.5 even exceeds that with the standalone GPT-4. 
A subsequent ablation study suggests that \name's two key components — length alignment and semantic alignment — are beneficial for improving consistency across different comparison forms.

\section{Background}
\label{sec:background}

\parh{Paradigms of Using LLM-based Evaluators.}~Recent work has explored using
LLMs such as GPT-4 to evaluate and compare the performance of AI
systems~\citep{wang2023large,chan2023chateval,zheng2023judging,hada2023large}.
Conceptually, there are two distinct LLM-based comparison paradigms:
\emph{standalone comparison} and \emph{pairwise comparison}. 
In the standalone comparison, LLM evaluators are provided with one answer at a time and are asked to score each answer independently. As a result, position bias is not an issue in standalone LLM evaluation and is therefore beyond the scope of this paper.
Nevertheless, we find that the absolute scores of LLM may lack clear interpretation.
To demonstrate this, we conducted a preliminary study where we examined the
consistency of standalone comparison across a total of 80 test cases, each
involving three sets of value ranges. 
Our findings indicate that the scores from standalone comparison do not strictly adhere to a linear mapping relationship across different scales (more discussion in~\AP~\ref{app:pre-study}).

Pairwise comparison presents two answers side-by-side and asks evaluators to
select the superior one. In particular, pairwise comparison methods can be
further categorized into three forms: \emph{score-based}, \emph{likert-based},
and \emph{relation-based}. In score-based comparison, evaluators assign a score 
to each answer and then compare these scores to determine the better answer. The likert-based
method~\citep{rajani2023llmlabels} requires evaluators to score answers on a 
likert scale~\citep{likert1932technique}, where lower scores indicate a strong
preference towards the first answer, middle scores represent a close tie, and
higher scores signal a preference for the second answer.
Additionally, the relation-based comparison solicits direct inputs from the 
evaluators about their preference for one answer over another.
This approach aims to avoid the use of potentially arbitrary scores,
guiding evaluators to make relative comparisons between answers instead.
The details of these three forms are shown in \AP~\ref{app:comparison-template}.

\parh{Position Bias in Pairwise Comparison.}~Despite the generally encouraging
performance of pairwise comparison methods, we note that LLM evaluators are not
perfect and can exhibit certain biases. A primary concern is the \textit{position
bias}~\citep{zheng2023judging,wang2023large}, whereby the LLM may favor the
first (or second) answer in a pairwise comparison, regardless of its content.
In fact, LLMs have shown notable sensitivity to small changes in
prompts~\citep{zhao2021calibrate,zhu2023promptbench}. For clarity, we provide
a formal definition of position bias as well as the consistency. As
illustrated in \T~\ref{tab:rela-template} (refer to \AP~\ref{app:comparison-template}), 
the evaluation input comprises a fixed template with three placeholders.
The input set for the LLM evaluators can be represented as
$\{Q, R_1, R_2\}$, where $Q$ denotes the question set, and $R_1$ and
$R_2$ are the two sets of answers for comparison. 
The LLM evaluators produce the verdict $V = LLM(\{Q, R_1, R_2\})$,
which indicates the preferred answer out of the two candidates. 
Assuming that the LLM evaluators are flawless, the verdict $V$ should be
independent of the permutation $\Pi$ of $R_1$ and $R_2$. Thus, position bias
can be expressed as: $ \Pi \not \! \perp\!\!\!\perp V$. On an individual sample level,
for a specific question $q \in Q$ and answers $r_1$, $r_2$,
consistency is achieved if the verdict $v$ remains the same when the positions of
$r_1$, $r_2$ are switched: $LLM(\{q, r_1, r_2\}) = LLM(\{q, r_2, r_1\})$.

\section{The \tool System}
\label{sec:method}

\parh{Design Intuition.}~It is worth noting that both human
evaluators and LLMs encounter difficulties in making consistent evaluations when
faced with lengthy and intricate
answers~\citep{KINTSCH1973257,wijesiriwardene2023analogical}. 
A common cognitive approach among individuals is to decompose information into smaller units,
thereby simplifying the comparison process~\citep{ratnasari2023students}. 
Inspired by this observation, \tool\ is
designed to split candidate answers into segments,
merge specific segments across candidates that share ``comparable'' content,
and eventually align them. Based on this intuition,
\name seeks to mimic effective human comparison procedures,
aiming to calibrate position bias and enhance the consistency of LLM evaluators.
That said, for each question, the verdicts of
\tool\ should be consistent with any permutation of the answers, i.e.,
-$LLM(\{q, r_1, r_2\}) = LLM(\{q, r_2, r_1\})$.

Moreover, \tool is specifically designed to address the position bias issue, rather than aligning the reasoning capabilities of LLM-based evaluators. Therefore, if an LLM evaluator lacks the capability to adequately judge a pair of semantically different responses, this falls outside the scope of \name's intended purpose. \name is solely focused on rectifying inconsistencies that arise when the order of a pair of semantically different responses is altered, leading to a change in the judgment of the LLM evaluator.
In this context, \tool is generalizable to handling open-ended answers because, regardless of their semantic differences, we can always mix the two previously individual answers. Such a mixing operation effectively eliminates the position bias in pairwise LLM-based evaluation.

\subsection{Key Design Considerations}
\label{subsec:key-feature}

Before presenting the technical details of \tool, we first
introduce its key design considerations.

\parh{Content Preservation.}~Content preservation refers to ensuring the
segmented answers encompass the entirety of the information present in the
original answer, without any omissions or additions of new content. For a given
original answer $r_1$, the set of split answer segments $\{r_1^{1}, r_1^{2},
..., r_1^{k}\}$ should fully encompass the content of $r_1$. This implies that
when the segments are concatenated, the entirety of the original content is
preserved ($\sum_{i=1}^{k} r_1^{i}= r_1$). 
This consideration helps to preserve the meaning and information of the original answer during the process of splitting.
The preservation of content is critical for
evaluators to assess the same substantive answer content that is divided into
segments, without any alterations or incomplete information.

\parh{Order Preservation.}~Order preservation refers to preserving the original
sequence of the information presented in the answer. This is important for fair
evaluation, as re-ordering or re-arranging the content may impact the assessment
of answer quality. 
By preserving the order, we ensure the
segmentation process does not introduce artifacts that could unintentionally
alter assessment. This enables the LLM evaluators to accurately evaluate
answers in comparison to the original. 
Notably, considering both the content and order of the answer helps maintain long-range dependencies by preserving all original information and the sequence in which it is presented.

\parh{Resource efficiency.}~Resource efficiency refers to minimizing
computational costs incurred by the splitting process, beyond the standard cost
when querying the LLM evaluator. To this end, it is important for the
segmentation process to introduce a minimal number of extra tokens and to be
executed rapidly, thus avoiding significant overhead.

\subsection{The Core Splitting Algorithm}
\label{subsec:algo}

Due to the page limit, we direct interested readers to \AP~\ref{app:overview} for a comprehensive overview of utilizing \name for LLM evaluation.
Here we concentrate on \name's core splitting algorithm, as illustrated in \A~\ref{alg:oracle}.
Intuitively, \tool\ first identifies semantically or
syntactically similar segments across answers. It then aligns these answer segments
and merges them sequentially into a single prompt for the LLM evaluators to make a final verdict.
Specifically, the inputs include the
question $q$, two candidate answers $r_1$ and $r_2$, the LLM evaluator's verdict
function $v()$, and the specified number of splits $k$. The output of
\A~\ref{alg:oracle} is a consistent verdict $v \in (1,2,3)$, where $1$ indicates
that $r_1$ is superior, $2$ suggests that $r_2$ is better, and $3$ represents a
tie.

Overall, the splitting process can be divided into three stages. In the first
phase, possible split positions are determined at the boundaries of sentences
(line 1-2). Segmenting at sentence breaks (e.g., periods or question marks) reduces the
likelihood of producing incomplete words or fragmented syntactic units in
different segments. This particular design decision aids in maintaining semantic
consistency and enhancing readability in each segment. 
Notably, natural language and programming language have different definitions
for sentence boundaries; for instance, the period sign ``.'' in Python denotes
accessing a specific object member property. Therefore, in instances where
answers involve code blocks, we follow~\citep{li2023protecting,wang2023reef} and leverage \texttt{treesitter}~\citep{treesitter}
to parse code blocks and locate suitable split positions that preserve the
code's structure and execution sequence.

\begin{algorithm}[!htbp]
    \footnotesize
    \caption{Alignment-based Splitting}
    \label{alg:oracle}
    \KwIn{Question: $q$, Answers: $r_1, r_2$, Evaluator's verdict $v()$, Split number $k$} 
    \KwOut{Consistent evaluation $v \in (1,2,3)$ }
    
    \tcc{Step1: identify answers' formats with split positions.}
    $r_1^{positions} = format(r_1)$\\
    $r_2^{positions} = format(r_2)$\\
    \tcc{Step2: length alignment.} 

    $[r_1^{(1)},...r_1^{(k)}] = equalsplit(r_1^{positions},k)$\\
    $[r_2^{(1)},...r_2^{(k)}] = equalsplit(r_2^{positions},k)$\\

    \If{$v(q_i, r_1^{(1)}, r_2^{(1)},...,r_1^{(k)}, r_2^{(k)}) == v(q_i, r_2^{(1)}, r_1^{(1)},...,r_2^{(k)}, r_1^{(k)})$}{
        
        \Return{v} \tcc{Consistent, return answer}
    }
    \tcc{Step3: semantic alignment.}
    \Else{
        $s_{max} = 0, n_{s}=0$, $Search\_all = False, r_1^{bestparts}=[], r_2^{bestparts}=[]$\\
        \While{not Search\_all}{
            $ r_1^{parts} = partition(r_1^{positions}, k, n_{s})$ \\
            $ r_2^{parts} = partition(r_2^{positions}, k, n_{s})$\\
            $n_{s} += 1$\\    
                $s_{cum} = \sum_{i=1}^{k} similarity(r_1^{parts}[i], r_2^{parts}[i])$\\
                \tcc{Update max similarity score, keep best split positions.}
                \If{$s_{cum} > s_{max}$}{
                    $s_{max} = s_{cum}, r_1^{bestparts}=r_1^{parts}, r_2^{bestparts}=r_2^{parts}$
                    }
            
        }
        \If{$v(q_i, r_1^{(1)}, r_2^{(1)},...,r_1^{(k)}, r_2^{(k)}) == v(q_i, r_2^{(1)}, r_1^{(1)},...,r_2^{(k)}, r_1^{(k)})$}{
        \Return{v}
        }
    }
    \Return{None}

\end{algorithm}

The second stage performs length alignment, splitting each answer into $k$
segments of comparable length (line 3-4); if an answer is too short to split, \tool\ would give up splitting and directly ask for the judge. Specifically, we first find the $k-1$
points that divide the answer into $k$ equal segments
according to the number of characters. Subsequently, we select the split
location that is closest to each of the split positions obtained in the first
stage, and designate them as $[r_1^{(1)},...r_1^{(k)}]$.\footnote{An
illustration with two detailed algorithms is available in~\AP~\ref{app:split-algo} to ease understanding.} The $k$ corresponding answer segments are subsequently merged again and
used for evaluation by the LLM evaluator. If the LLM evaluator consistently
returns the same verdicts for all length-aligned splits, then the verdict is
returned (lines 5-7).

If inconsistent assessments persist after length alignment, \tool\ proceeds to
semantic alignment as the third stage (lines 8-17). Specifically, given a fixed
$k$ and a set of possible split positions, we aim to iteratively
search for the optimal split positions that maximize the cumulative semantic
similarity between corresponding segments of the two answers. 
Note that $n_s$ represents the index number of the current segmentation, and $Search\_all$ becomes \texttt{True} when $n_s$ reaches the maximum number of possible split combinations $Cal$.
Semantic similarity between segments
$r_1^{t}$ and $r_2^{t}$ is computed by token overlap: $sim\_score =
\frac{Intersection(set(r_1^{t}),set(r_2^{t}))}{\max(\text{len}(set(r_1^{t})),\text{len}(set(r_2^{t})))}$.
Notably, the choice of value $k$ as well as the similarity metric would have an
impact on the efficiency of \tool, and we provide the theoretical analysis in
\S~\ref{subsec:efficiency}. 
We also consider applying other similarity metrics, such as LM-based
metrics~\citep{reimers-2019-sentence-bert}. However, we argue that employing such intricate metrics is
not necessary for \tool, as they usually entail extra computing resources, and
introduce more hyper-parameters while yielding only marginal improvements in
performance; see further discussion in~\AP~\ref{app:lm-metric}. 
Finally, \tool\ would yield consistent verdict if applicable (lines 19-22).
Note that the above three stages are carried out in a sequential manner, whereas
semantic alignment is only performed when 
length alignment is inadequate for ensuring consistent assessments. 
This sequential approach prioritizes computational efficiency, as length alignment is typically faster to execute than semantic alignment.

\section{Experiments}
\label{sec:evaluation}

\subsection{Experimental Setup}
\label{subsec:settings}

\parh{Datasets.}~We evaluate \tool\ using the MT-Bench
benchmark~\citep{zheng2023judging}, following the experimental setup in
\cite{wang2023large}. MT-Bench contains 80 elaborated open-ended questions spanning 8
categories (Writing, Roleplay, Reasoning, Math, Coding, Extraction, STEM,
and Humanities).
For each question, MT-Bench provides several candidate answers from different
LLMs. We consider eight different combinations of LLM answers (see more details
in~\AP~\ref{app:llm-details}), and we consider all three comparison forms
(score-based, likert-based, and relation-based) in the pairwise comparison
paradigm. Thus, we have $80*8*3=1920$ inputs to evaluate
each LLM evaluator. 
We use this diverse dataset to provide a comprehensive evaluation of \tool\ across several representative LLMs and comparison forms. Additionally, an extended evaluation on a larger set of open-ended questions can be found in~\AP~\ref{app:generalizability}.

\parh{Models.}~In this work, we include both locally deployable
models that are open-source and proprietary models that are accessed through
only cloud APIs as LLM evaluators. 
Details on the specific LLM versions evaluated are given in~\AP~\ref{app:llm-details}.

\parh{Response Length and generalizability.}~
In \F~\ref{fig:example}, two assistants provide responses of similar length. However, given the open-ended nature of the questions in MT-Bench, different LLMs may produce responses that differ substantially in both length and content for the same question. To assess \tool's adaptability to open-ended questions, we analyzed the statistics of all responses, presented in \T~\ref{tab:response-length-statistic}. 
Our findings indicate that the lengths of responses from the LLMs vary considerably, underscoring \tool's flexibility in handling open-ended questions. 
Additionally, we explore \tool's generalizability by examining the relationship between answer length and inconsistency (\AP~\ref{appsub:response-len-inconrate}) and evaluate its performance on extremely short responses (\AP~\ref{appsub:response-len-short-extreme}).
More details are provided in~\AP~\ref{app:response-len}.

\begin{table*}[!htbp]
    \small
    \footnotesize
    \centering
    \begin{tabular}{c | c | c | c | c | c}
        \toprule

        \textbf{Evaluators}& \textbf{De. Method}  & \textbf{Model} & \textbf{Relation-based} & \textbf{Score-based} & \textbf{Likert-based} \\
        
        \midrule

        \multirow{3}*{Claude2} 
        & &\% Origin Con & 28.28 & 47.34 & 50.62   \\
        & API&\% \tool\ Con
        & \textbf{83.28(+194.48\%)}   
        & \textbf{65.16(+37.64\%)} 
        & \textbf{94.84(+87.36\%)}\\
        & &\% Fixed Coverage & 79.44 & 52.22 & 91.27   \\

        \midrule

        \multirow{3}*{Qwen} 
        & &\% Origin Con & 63.12 & 52.66 & 8.12    \\
        & API &\% \tool\ Con
        & \textbf{78.13(+23.78\%)} 
        & \textbf{71.09(+35.0\%)} 
        & \textbf{9.38(+15.52\%)} \\
        & &\% Fixed Coverage & 65.66 & 59.78 & 6.46   \\
        \midrule

        \multirow{3}*{Chatglm2} 
        & &\% Origin Con & 38.44 & 58.59 & 26.72  \\
        & Local&\% \tool\ Con
        & \textbf{61.72(+60.56\%)} 
        & \textbf{74.06(+26.4\%)} 
        & \textbf{64.22(+140.34\%)} \\
        & &\% Fixed Coverage & 56.09 & 51.02 & 60.30   \\
        \midrule

        \multirow{3}*{Llama2} 
        & &\% Origin Con & 36.41 & N/A & N/A   \\
        & Local&\% \tool\ Con
        & \textbf{68.75(+88.82\%)}  
        & \textbf{N/A} 
        & \textbf{N/A} \\
        & &\% Fixed Coverage & 22.51 & N/A & N/A  \\
        \midrule

        \multirow{3}*{GPT-3.5} 
        &  & \% Origin Con  & 78.12 & 39.22 & 78.91 \\
        & API & \% \tool\ Con
        & \textbf{88.59(+13.4\%)} 
        & \textbf{54.84(+39.83\%)} 
        & \textbf{98.60(+24.94\%)} \\
        & & \% Fixed Coverage  & 70.63 & 42.06 & 96.32  \\
        \midrule

        \multirow{3}*{GPT-4} 
        & &\% Origin Con & 93.44 & 92.75 & 61.50   \\
        & API &\% \tool\ Con
        & \textbf{97.03(+3.84\%)} 
        & \textbf{98.00(+5.66\%)}
        & \textbf{63.50(+3.25\%)}  \\
        & &\% Fixed Coverage & 80.99 & 86.33 &  36.09  \\
        \bottomrule
    \end{tabular}
    \caption{The main results of \tool\ across LLM evaluators. All metrics
    presented are higher-is-better values. ``\% Origin Con'' and ``\% \tool\
    Con''  are the percentages of consistent results in the original setting
    when enhanced by \tool, respectively. ``\% Fixed Coverage'' denotes the
    percentage of inconsistent original assessments that are later corrected by
    \tool. ``De Method'' specifies whether the LLM evaluator uses local or cloud
    API deployment.}
    \label{tab:main-result}
\end{table*}

\subsection{Main Results}
\label{subsec:main-results}

As shown in \T~\ref{tab:main-result}, \tool\ improves the consistent rate among
all evaluators. The values depicted in the table correspond to the mean values
obtained from the analysis of all eight combinations of tested models. We
observe that \tool\ relatively improves the consistent rate by 3.25\% to
194.48\%, depending on the evaluator, with the highest fixed coverage at 96.32\%
(meaning that nearly all the inconsistent results are resolved). GPT-4 exhibits
the highest average consistency rate, which is in line with the findings of
previous work~\citep{wang2023large}, and \tool\ further boosts its consistency
up to 98\%. Moreover, we observe that GPT-4 exhibits subpar performance
on the likert-based form, not just compared to its performance on other forms,
but also when compared to GPT-3.5. Upon analyzing results on likert-based forms,
over 78\% of GPT-4's inconsistency provides a score of 5, reflecting its bias
for the second answer, and our method rectifies 36.09\% of them. Notably, we
only report the results of Llama2 in relation-based form, as it fails to provide
meaningful evaluations in other forms (see more details
in~\AP~\ref{app:bad-example}).

The impact of the comparison form on consistency rates is also observed, with
evaluators displaying various preferences. For instance, it is seen that GPT-3.5
exhibits the least consistent performance when evaluated on the score-based
form, whereas Claude2 struggles most on the relation-based form. GPT-4, Qwen,
and Chatglm2 exhibit the highest degree of inconsistency when assessed on the
likert-based form. This suggests that appropriately matching comparison forms to
evaluators' capabilities is important. Nevertheless, \tool\ offers high
enhancement for forms and LLM evaluators. 

\parh{Per-category Breakdown Analysis.}
We recognize that our method may not perfectly resolve all inconsistencies, and there might be instances where it fails. To better understand \tool's performance across different categories, we conduct a statistical analysis of its failure rates across 8 categories, as shown in~\T~\ref{tab:failure-rates}.

\begin{table}[h]
    \centering
    \resizebox{1.0\linewidth}{!}{
    \begin{tabular}{c|c|c|c}
    \hline
    Category & FR & Category & FR \\
    \hline
    Coding & 17.13\% & Common-sense & 12.15\% \\
    Fermi & 14.92\% & Math & 11.60\% \\
    Counterfactual & 13.81\% & Knowledge & 9.94\% \\
    Generic & 12.15\% & Roleplay & 8.29\% \\
    \hline
    \end{tabular}
    }
    \caption{Failure Rates (FR) of \tool\ across different categories.}
    \label{tab:failure-rates}
\end{table}

We observe that \tool\ tends to fail more frequently on coding-related questions. This may be attributed to the tighter contextual relationships in code due to programming language constraints and fixed syntactic structures (e.g., for loops). Although we have considered parsing code and designing separate split positions during the split stage (as shown in~\S~\ref{subsec:algo}), these results indicate room for improvement in handling structured data.

To further investigate the generalizability of \tool, we conduct additional experiments on more open-ended questions, and the results are provided in~\AP~\ref{app:generalizability}.
Additionally, we compare \tool\ with more baselines, where results are shown in~\AP~\ref{app:stronger-baseline}.

\subsection{Efficiency and Cost Analysis}
\label{subsec:efficiency}

\begin{figure*}[!htbp]
    \centering
    \begin{subfigure}[b]{0.49\textwidth}
        \includegraphics[width=\textwidth]{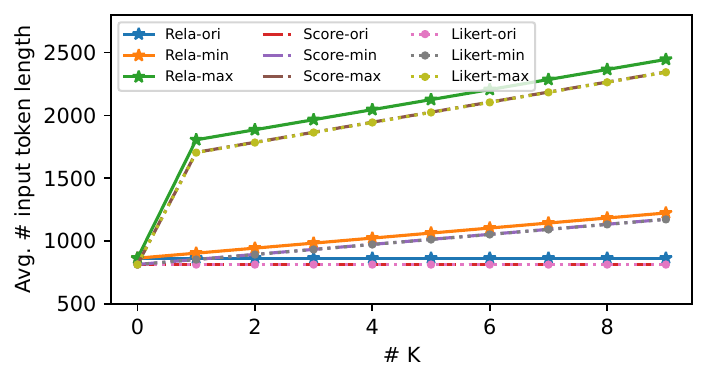}
        \caption{Average input token length with different $k$. ``ori''
        represents the original input length. ``min'' and ``max'' represent the
        minimum and maximum input lengths, respectively.}
    
    \end{subfigure}
    \hfill
    \begin{subfigure}[b]{0.49\textwidth}
        \includegraphics[width=\textwidth]{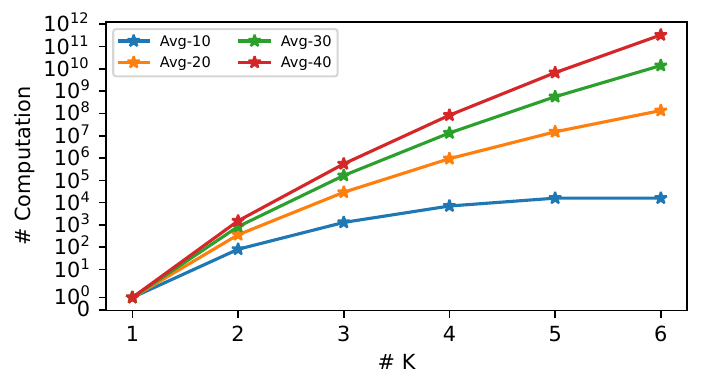}
        \caption{Computation operations with different $k$. The number behind
        ``Avg'' is the average potential split positions for each answer. The
        y-axis has been logarized to ease reading.}
    \end{subfigure}
    \caption{Theoretical estimation of \name's cost with varying $k$ based on input length and computation operations.}
    \label{fig:efficiency}
\end{figure*}

To show the efficiency and cost-effectiveness of \name,
this section first performs a theoretical analysis of \name's efficiency,
and then evaluates its actual costs in terms of temporal, monetary, and
environmental factors. Specifically, we measure the efficacy of \tool-enhanced LLMs in terms of their agreement rate with GPT-4.

\parh{Theoretical analysis.}~We first theoretically analyze \tool's computational efficiency.
As the number of answer segments $k$ increases, the average input
length for LLM evaluators also grows correspondingly. In line with line 1
in~\A~\ref{alg:oracle}, the added tokens stem from two sources: fixed-length
system prompts based on comparison forms, and split boundary prompts (an example
shown in~\T~\ref{app:alignment-template}) that scale linearly with $k$. Consequently,
the additional input length scales as $O(K)$, as depicted
in~\F~\ref{fig:efficiency}(a). Notably, the average input length of the relation-based form exceeds the other two, as it
requires more tokens for the system prompt. More details are in~\AP~\ref{app:prompt-template}.

In accordance with line 9 in~\A~\ref{alg:oracle}, the total number of
computation operations $Cal$ is calculated as: $Cal = C_{p_1}^{k-1} *
C_{p_2}^{k-1} $, where $p_1$ and $p_2$ are the potential split positions in the two
answers. $C_{p_1}^{k-1}$ and $C_{p_2}^{k-1}$ are the combination counts
for the first and second answers, respectively. Using average position numbers of
10, 20, 30, and 40, we derive the total calculations as depicted
in~\F~\ref{fig:efficiency}(b). 
Intuitively, raising the value of $k$ can improve the algorithm's performance
by exploring more split position combinations. However, this also
results in an exponential surge in the total computation operations,
compromising efficiency. As such, we conducted controlled experiments to
identify the optimal value of $k$, and in our case, we found that setting $k=3$
strikes a balance between efficiency and precision. Full details about this
controlled experiment can be found in~\AP~\ref{app:lm-metric}.

\parh{Real-World Performance and Cost Analysis.}~Next, we measure the level of
agreement between the \tool-enhanced LLM evaluators and 
GPT-4 (considered as the ``reference standard.''). Note that to offer a
fair evaluation, we exclusively consider GPT-4 evaluation outputs that are
originally consistent. In the context of a question with two possible answers,
it is deemed as an agreement only when both GPT-4 and \tool-enhanced assessments
are consistent and identical. As evidenced in~\T~\ref{tab:agreement-cost},
agreement rates are enhanced by an average of 16.32\% after alignment. Claude2
has the highest gain at 31.65\%, while GPT-3.5 achieves the highest agreement
rate with GPT-4 at 88.59\%.

Additionally, we consider the resource usage in terms of temporal,
monetary, and environmental factors.
As shown in~\T~\ref{tab:agreement-cost}, Chatglm2 exhibits the lowest inferencing time. 
However, the cost of GPT-3.5 is lower than that of Chatglm2, while its
carbon emission is higher, which is mainly because
the cloud API models usually run on GPU clusters with more powerful GPUs. 
It is worth mentioning that GPT-3.5 incurs less than \textbf{10\%} of the average
cost of GPT-4, while maintaining an approximate agreement level of 88\% with
GPT-4. In brief, the usage of \tool\ results in a substantial level of
concurrence with GPT-4 while maintaining a minimal computational burden, hence
showcasing a proficient and eco-friendly alignment. The significant enhancements
in performance and resource utilization underscore the usefulness of this
approach in boosting various LLMs for crucial evaluation work.

\subsection{Human study}
\label{subsec:human-study}

We conducted a human evaluation to further assess the performance of \tool.
The model pair ``gpt-3.5-turbo'' v.s. ``Claude-v1'' is selected to compare human agreement rates on original versus \tool-enhanced assessments across 80 questions, as these two models have similar performance~\citep{zheng2023judging}, making it challenging for LLM evaluators to make decisions.
We recruit five experts: two industrial developers and three academic researchers, none of whom are authors of this paper to avoid potential bias due to prior exposure to the MT-BENCH dataset.
For each participant, we create an online questionnaire that provides one question with two answers,
not specifying their origin. Before the questionnaire, brief instructions on the
task and evaluation criteria are provided. (More details are in~\AP~\ref{app:annotation-process}.)
During the human evaluation process, we observe instances where human evaluators have differing assessments. This aligns with previous research highlighting the diversity of human perspectives~\citep{peng1997validity}. 
In such cases, we employ a majority vote to determine the final result, and we aim to use \tool\ to help LLM judges closely align with representative human evaluations.

\begin{table}[!htpb] 
    \centering 
    \footnotesize
    \begin{tabular}{l|c|c} 
        \hline
        & Ori HAR (\%) & Fix HAR (\%) \\  \hline
        GPT-3.5 & 55.00 & 63.75 \\
        Qwen & 35.00 & 35.00 \\
        Chatglm2 & 16.25 & 17.50 \\
        Claude2 & 6.25 & 47.50 \\ 
        GPT-4 & 60.00 & 65.00 \\ \hline
    \end{tabular}
    \caption{Main results from human evaluation comparing the model pair ``gpt-3.5-turbo'' v.s. ``Claude-v1'' on 80 questions. ``HAR'' represents the human agreement rate.}
    \label{tab:humaneval2}
\end{table}

The human evaluation results presented in~\T~\ref{tab:humaneval2} demonstrate
increased agreement rates between humans and LLM evaluators after applying
\tool. For example, the human agreement rate with GPT-3.5 increases from 55.00\% on the original assessments to 63.75\% after applying \tool\ enhancements, which surpasses the original human agreement rate with GPT-4.
In addition, the original human agreement rate for Claude2 is only 6.25\%, but increases substantially
to 47.50\% after enhancement. 
Taken together, these quantitative findings provide evidence that \tool\ effectively
augments the assessments of all LLM evaluators to achieve greater concordance
with human evaluators. The framework also enables weaker LLMs to reach
comparability with stronger counterparts in terms of human alignment. 

It is crucial to recognize that while LLM judges offer significant benefits in terms of cost and efficiency, they do not entirely replace the need for human expertise. By thoroughly investigating the potential biases present in LLM judges and understanding their limitations across different categories, we can develop more effective and comprehensive methods for evaluating AI systems. This improvement not only enhances the quality and efficiency of assessments but also brings substantial benefits to the entire AI community.

\begin{figure*}[!htpb]
    \centering
    \includegraphics[width=0.85\textwidth]{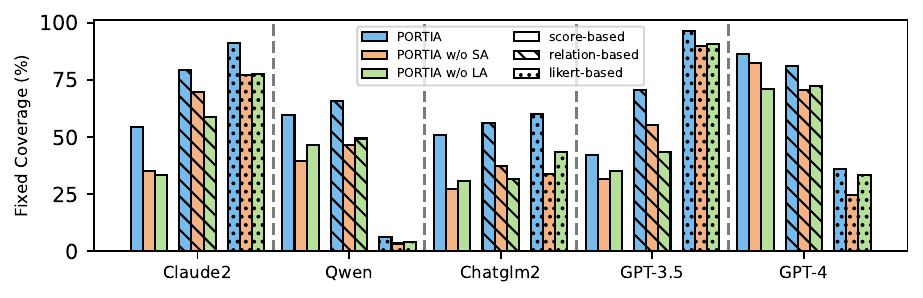} 
    \caption{Fixed coverage rate across LLMs for \tool\ and variants w/o Semantic (SA) or Length Alignment (LA).}
    \label{fig:ablation}
\end{figure*}

\subsection{Ablation Study}

To ascertain the individual contributions of each component in \tool, we conduct
ablation experiments on five distinct LLM evaluators. The results are quantified
in terms of the ``Fixed Coverage'' metric, as depicted in~\F~\ref{fig:ablation}.
To facilitate visual interpretation,
variants of \tool\ incorporating or excluding specific components are denoted by
different colored bars in the histogram. Additionally, texture patterns in the
bars indicate the comparison form used. The plain blue bar represents the
score-based form, while the blue bar with slash lines corresponds to the
relation-based form.

The results reveal that both semantic and length alignment improve \tool's performance.
Specifically, across all evaluators, semantic alignment shows a greater contribution to enhancing the likert-based form, possibly because the likert scale has a greater dependence on precise semantic meaning for its standardized categorical ratings.
For the other two forms, both alignment methods have similar contributions with slight differences between LLMs. This is likely because these forms consider semantic meaning and answer length in a balanced way, similar to how human evaluators would~\cite{ratnasari2023students}.

Furthermore, we find that the trends of fixed coverage rate are consistent across comparison forms for \tool\ and ablations (without semantic or length alignment).
Likert-based form has the highest fixed coverage rates, followed by relation-based, with score-based form having the lowest rates.
The exceptions are Qwen on likert-based form and GPT-4 on all forms, where we manually check and find that: 
(1) Qwen prefers the second answer for over 90\% of examples, no matter whether \tool\ is used.
(2) GPT-4 has the highest fixed coverage rate on relation-based form, which is probably because GPT-4 performs consistently enough (more than 97\% consistent rate), 
and therefore, the improvement on likert-based form is not obvious.
Overall, aside from the outliers, these results suggest that likert-based form is the most challenging, and we attribute this to that it requires the evaluators to assign a single score that contains an assessment of two answers, which is more difficult than simply choosing the better one like relation-based form.

\section{Related Work}
\label{sec:related}

\parh{Automatic Evaluation of AI Outputs.}~Automated evaluation metrics like BLEU~\citep{papineni2002bleu} and ROUGE~\citep{lin2004rouge} are widely used to assess the quality of AI-generated text. However, these metrics have limited ability to assess meaning, reasoning, and logical consistency. Recent efforts have focused on developing more robust semantic similarity metrics using neural representations~\citep{zhang2019bertscore}, but they are still imperfect proxies for human assessment. To address this, LLM has emerged as a promising alternative for evaluation~\cite{chiang2023large,liu2023geval,Huang2023EmotionallyNO,Jiao2023IsCA,lin-chen-2023-llm,wang2023pandalm}.

\parh{Biases in LLM Evaluators.}
Besides position bias, Zheng et al.~\citep{zheng2023judging} identify two additional biases: verbosity bias, which refers to a preference for longer answers, and self-enhancement bias, which involves a preference for self-generated answers. 
However, the definition of verbosity bias remains ambiguous, and in line with previous research~\citep{wang2023large}, we observe that human evaluators also tend to prefer longer answers. 
Furthermore, self-enhancement bias is not universal for all LLMs~\citep{zheng2023judging}. 
Therefore, we focus on position bias, as its mitigation can directly improve the efficiency and accuracy of various LLM evaluators already in real-world use~\citep{alpaca_eval}.

\section{Conclusion}
\label{sec:conclusion}

This paper presented \tool, an alignment-based technique to address position
bias for LLM evaluators. By aligning similar content segments across candidate answers,
\name effectively reduced position bias.
It not only enabled replacing costly models like GPT-4 with affordable alternatives but also elevated the consistency rate of open-source models like Llama2.

\section{Acknowledgements}
\label{sec:acknowledgements}

The HKUST authors are supported in part by a RGC GRF grant under the contract 16214723, RGC CRF grant under the contract C6015-23G, research fund provided by HSBC, and a Webank research fund WEB24EG01. 
The HITSZ authors are supported in part by National Natural Science Foundation of China under project (No. 62472126), Natural Science Foundation of Guangdong Province (Project No. 2023A1515011959), Shenzhen-Hong Kong Jointly Funded Project (Category A, No. SGDX20230116091246007), and Shenzhen Basic Research (General Project No. JCYJ20220531095214031)
We are grateful to the anonymous reviewers for their valuable comments.

\section{Ethical Considerations}
\label{sec:ethics-stmt}

\parh{Use of Human Annotations}~We protect the privacy rights of workers and pay them above the local minimum wage. All five annotators are paid 35\$\ per hour for the given 80 samples. Careful instruction is given to ensure that the annotators understand the task and are not exposed to harmful content.

\parh{Study Scope.}~Our work aims to improve the consistency of LLM-based evaluators, which can be utilized to assess the quality of AI-generated answers. More consistent LLM-based evaluators can provide human-like evaluations at a lower cost, supplying feedback to reduce biases during training. 
Notably, our work enhances the evaluation consistency of open-source models like Llama2, enabling their use as reliable evaluators and making research on LLMs more accessible to the broader community.
However, we recognize that malicious actors could exploit these methods to intentionally train models that go against human values. 
The open-source LLMs could be leveraged as consistent evaluators to guide the training of harmful models such as Worm-GPT~\citep{wormgpt}. 
While our work targets constructive applications, we caution that like any technology, consistent LLM evaluators could potentially be misused. Researchers should consider ethical implications and preventative measures. 
Overall, our current focus is on addressing the position bias of LLM evaluators, thereby making them more consistent and reliable in supporting large-scale automatic evaluation processes.

\section{Limitations}
\label{sec:limitations}

\parh{Context Window Length.}~Although we do not identify any new biases introduced by \tool\, the maximum context window length of the LLM evaluator poses a challenge, as it requires the LLM to process the entire input prompt, which comprises the original question and two candidate responses. If these responses are excessively lengthy, the merged prompt may exceed the maximum context window length of the LLM, thereby violating the crucial principle of content preservation discussed in \S~\ref{subsec:key-feature}. While we did not encounter this issue in our experiments, it could potentially limit the effectiveness of LLMs with shorter context window lengths. We believe that this can be addressed by increasing the maximum context window length of LLMs~\cite{xiao2023efficient} or simply selecting an LLM with a longer context window length, such as Claude2 for 100k tokens.

\parh{Excessive LLM alignment.}~Although \tool\ works well under most scenarios,
it is not perfect. One limitation is that \tool\ is not able to handle the case
where the LLM evaluators refuse to make a verdict, which usually occurs on LLMs
with advanced alignment techniques such as GPT series models. We find that these models would become too conservative to make a verdict, no matter how the answers are split and aligned. For example, GPT-3.5 often refuses to give any meaningful response when the question is in ``Roleplay'' category.

\bibliography{custom,bib/llm,bib/cot,bib/sast,bib/others,bib/zj}

\newpage
\newpage
\appendix

\section{Reproducibility}
\label{app:reproducibility}

To assure reproducibility, we employ various
methods to mitigate the inherent randomness in the decoding process of LLMs. For
models using cloud API, the hyper-parameter ``temperature'' is uniformly set to
0 across all evaluators. For local models, the sampling function is deactivated
during the decoding phase to get deterministic results.
Specifically, we run experiments on a GPU server with Intel Xeon
Platinum 8276 CPU, 256GB of RAM, and 4 NVIDIA A100 GPUs. This server is capable
of performing cloud API calls and local LLM inference.

All our results are reproducible using the code repository we will
release. All experimental details, including hyperparameters, are reported in~\S~\ref{subsec:algo} and~\AP~\ref{app:lm-metric}.
We reuse the benchmark datasets from~\cite{zheng2023judging}, with the different comparison prompt forms detailed in~\AP~\ref{app:prompt-template}. 
\section{Response Length}
\label{app:response-len}

\subsection{Response Length Statistics}
\label{appsub:response-len-stat}

It is possible for the generated results to differ significantly from each other.
To further explore this, we conducted an analysis of the statistical information of all LLM responses, revealing substantial differences in response lengths within our benchmark dataset. The relevant data is presented in~\T~\ref{tab:response-length-statistic}.

\begin{table*}[!htpb] 
    \centering 
    \footnotesize
    \begin{tabular}{c|c|c|c|c} \hline 
        LLM & Max Length & Min Length & Average Length & Standard Deviation \\ \hline 
        Alpaca-13b & 1149 & 6 & 508.99 & 222.98 \\
        Bard & 2652 & 151 & 1276.62 & 495.88 \\
        Vicuna-7b & 2598 & 266 & 1457.01 & 448.55 \\ 
        Claude-v1 & 2392 & 94 & 1624.8 & 612.29 \\ 
        GPT-3.5 & 2218 & 193 & 1206.29 & 460.26 \\ 
        Vicuna-13b & 2441 & 212 & 1416.92 & 371.0 \\ 
        GPT-4 & 3842 & 201 & 2044.14 & 768.59 \\ 
        Llama-13b & 4827 & 9 & 757.57 & 895.74 \\ \hline
        GPT-3.5-short & 365 & 26 & 152.55 & 58.07 \\ \hline
    \end{tabular}
    \caption{The statistics of answers from different LLMs. ``GPT-3.5-short'' is generated by instructing GPT-3.5 to shorten its responses while preserving as much meaning as possible, which consists of responses approximately 1/8th the length of the original ones.}
    \label{tab:response-length-statistic}
\end{table*}

From the table, we observe that the lengths of responses generated by the LLMs vary considerably. For example, the maximum number of characters in the responses is 4,827, generated by Llama-13b, while the minimum is just 6 characters, generated by Alpaca-13b.

\subsection{Relationship Between Answer Length and Inconsistency}
\label{appsub:response-len-inconrate}

To further explore the relationship between answer length and inconsistency, we conduct an additional experiment using the collected judgment data. For this, "GPT-3.5" was used as the evaluator, analyzing 8 pairs of responses across three comparison forms.
The answers are categorized into 9 groups based on their length, with each group representing an 800-character interval. The resulting data are presented in \T~\ref{tab:char-range-incon}, with values below 2\% of the total indicated by ``-''.

\begin{table}[!htpb] 
    \footnotesize
    \centering 
    \begin{tabular}{c|c} \hline 
        Char Range (*100) & \% Incon Rate \\
         \hline 0-8 & - \\ \hline 8-16 & - \\ \hline 16-24 & 26.89 \\ \hline 24-32 & 23.02 \\ \hline 32-40 & 31.84 \\ \hline 40-48 & 39.01 \\ \hline 48-56 & 42.73 \\ \hline 56-64 & 55.45 \\ \hline 64+ & - \\ \hline 
        \end{tabular} 
        \caption{The inconsistency rates in different character count gaps.}
    \label{tab:char-range-incon}
\end{table}

The table shows a generally positive correlation between answer length and inconsistency rate, with shorter answers tending to exhibit lower inconsistency rates. This finding suggests that position bias is less significant in shorter answers. 
To clarify, \tool\ is designed to be adaptable to open-ended questions and answers without making any assumptions about the content of the candidate answers. As long as the responses contain sufficient content (at least two sentences in our current setup) for splitting, \tool\ will follow the same process to first split the responses and then conduct length or semantic alignment to merge them.
When combined with the enhancement results detailed in~\S~\ref{subsec:main-results}, this leads to the conclusion that the proposed framework is effective in handling responses of varying lengths.

\subsection{Extremely Short Response}
\label{appsub:response-len-short-extreme}

It is worth noting that our initial considerations did not account for a scenario where responses from one specific LLM are consistently and significantly shorter (e.g., 1/8th the length) than those from another. This is due to the expectation that LLMs under test are trained to generate responses adhering to given instructions, typically resulting in average response lengths of several hundred characters.

To determine whether our framework remains applicable in such special cases, we conducted an additional experiment with the following steps: (1) We instructed GPT-3.5 to shorten its responses while preserving as much meaning as possible, leading to a subset termed ``GPT-3.5-short,'' which consisted of responses approximately 1/8th the length of the original ones. (2) We then used GPT-3.5 and GPT-4 as evaluators to compare ``GPT-3.5-short'' with ``GPT-3.5'' and ``Claude-v1'' in exchanged orders, to assess consistency.

The results, shown in~\T~\ref{tab:short-res-consistency}, indicate a 100\% consistency rate (80/80) for both GPT-3.5 and GPT-4 as evaluators. This suggests that there is no inconsistency in this particular scenario, and therefore no alignment is needed. It means that position bias is no longer a concern in such situations. This finding aligns with previous studies~\cite{chiang2023large,liu2023gpteval}, which noted that LLM-based evaluators tend to assign higher scores to longer responses.

\begin{table}[h] 
    \footnotesize
    \centering 
    \begin{tabular}{c c|c|c} \hline 
    \multicolumn{2}{c|}{Evaluators} & GPT-3.5 & GPT-4 \\ \hline 
    Model1 & Model2 & & \\ \hline 
    GPT-3.5-short & GPT-3.5 & 100\% & 100\% \\ \hline 
    GPT-3.5-short & Claude-v1 & 100\% & 100\% \\ \hline 
    \end{tabular} 
    \caption{The consistency rates of GPT-3.5 and GPT-4 as evaluators for extremely short responses.}
    \label{tab:short-res-consistency}
\end{table}

It is worth noting that we have previously addressed the scenario in which the content of the responses differs significantly, but the length remains similar, as discussed in \S~\ref{subsec:main-results}. In the current case, the content of the responses is indeed different, as each response contains only one-eighth of the original content. However, there is no requirement for alignment in this particular scenario, as position bias is not a pertinent concern.
To summarize, we argue that \tool\ is applicable to open-ended questions, accommodating responses that vary significantly from each other, even in cases where one set of responses is systematically and markedly shorter than the other.

\subsection{Relationship Between Response Length Gap and Fixed Coverage}
\label{appsub:response-len-short}

To further explore the relationship between the gap in length between responses
and fixed coverage rate, we conducted an experiment using the collected judgment
data. For this, ``GPT-3.5'' was used as the evaluator, analyzing 8 pairs of
responses across three comparison forms.

\begin{table}[h] 
    \footnotesize
    \centering 
    \begin{tabular}{c|c|c} \hline & \% Fixed coverage & \% Frequency \\ \hline      
    0-300 & 50.82 & 0.37 \\ \hline 
    300-600 & 48.41 & 0.24 \\ \hline 
    600-900 & 63.3 & 0.17 \\ \hline 
    900-1200 & 62.67 & 0.11 \\ \hline 
    1200-1500 & 69.77 & 0.08 \\ \hline 
    \end{tabular}
\caption{Fixed coverage rates in different character count gaps.} 
\label{tab:char-gap-fixed} 
\end{table}

The answers are categorized into 5 groups based on their length, with each group
representing a 300-character interval. The results are presented in~\T~\ref{tab:char-gap-fixed}, with
frequencies below 3\% of the total being disregarded.

\section{Naming Reason}
\label{app:name-reason}

The name \tool is inspired by the intelligent and astute character, Portia, from Shakespeare's ``The Merchant of Venice.'' In the play, Portia assists a judge in making fair decisions within the legal rules. Just as Portia requests the exact amount of flesh to be cut, our method seeks to make fair splits of the original answers for comparison.
\section{A Preliminary Study of Standalone Comparison}
\label{app:pre-study}

\begin{table*}[hbp]
    \small
    \begin{tcolorbox}
    
    [Question] \textcolor[rgb]{0,0,0.9}{\{Q\}}
    
    [The Start of Assistant A's response] \textcolor[rgb]{0,0,0.9}{\{R1\}} [The End of Assistant A's response]
    
    [The Start of Assistant B's response] \textcolor[rgb]{0,0,0.9}{\{R2\}} [The End of Assistant B's response] 
    
    [System] 
    
    We would like to request your feedback on the performance of two AI assistants in response to the user question displayed above.
    
    Please rate the helpfulness, relevance, accuracy, level of details of their responses. Each assistant receives an overall score on a scale of 1 to 10, where a higher score indicates better overall performance.
    
    Please first output a single line containing only two values indicating the scores for Assistant A and B, respectively. The two scores are separated by a space. In the subsequent line, please provide a comprehensive explanation of your evaluation, avoiding any potential bias and ensuring that the order in which the responses were presented does not affect your judgment.

    We would like to request your feedback on the performance of one AI assistants in response to the user question displayed above.
    
    Please rate the helpfulness, relevance, accuracy, level of details of their responses. The assistant receives an overall score on a scale of \textcolor[rgb]{0,0,0.9}{\{min\_score\}} to \textcolor[rgb]{0,0,0.9}{\{max\_score\}} (with a minimum interval of \textcolor[rgb]{0,0,0.9}{\{interval\}}), where a higher score indicates better overall performance.
    
    Please first output a single line containing only one value indicating the score for Assistant. In the subsequent line, please provide a comprehensive explanation of your evaluation, avoiding any potential bias and ensuring that the order in which the responses were presented does not affect your judgment.

\end{tcolorbox}
\caption{The score-based evaluation form for standalone comparison with six slots (\textcolor[rgb]{0,0,0.9}{\{Q\}, \{R1\}, \{R2\}, \{min\_score\}, \{max\_score\}, \{interval\} }).}
\label{tab:score-template-single}
\end{table*}

In this section, following the same setting as~\cite{zheng2023judging}, we
conduct a preliminary study of standalone score-based LLM comparison.
We use the template shown in Table~\ref{tab:score-template-single} to generate
the input for LLM evaluators. For each question, we generate three sets of value
ranges, setting min\_score to 0, max\_score to 1, 10, and 100, and interval to
0.1, 1, and 10, respectively. In theory, if the standalone comparison answer is steady and
robust, the score should scale accordingly to the value ranges. For example, if
the score is 0.7 when the max\_score is 1, the score should be 7 when the
max\_score is 10, and 70 when max\_score is 100.

The LLM evaluators are asked to score each answer independently. We use the
answers from ``llama-13b'' as the input for LLM evaluators, and choose
``GPT-3.5'' as the LLM evaluator. Among a total of 80 test cases, we find that
the standalone comparison does not remain consistent for any of them.
Therefore, we conclude that the absolute scores of standalone comparison do not strictly adhere to a linear mapping relationship across different scales, potentially undermining their significance.
It is worth noting that although standalone comparison has been used in prior research by~\cite{chiang2023large,liu2023geval,zheng2023judging} to evaluate open-ended questions. It does not involve comparing two responses together, thereby eliminating any position bias. As a result, our paper primarily focuses on the position bias in pairwise comparison.

\section{\name's Pipeline}
\label{app:overview}

\begin{figure*}[!htpb]
    \centering
    \includegraphics[width=1.0\textwidth]{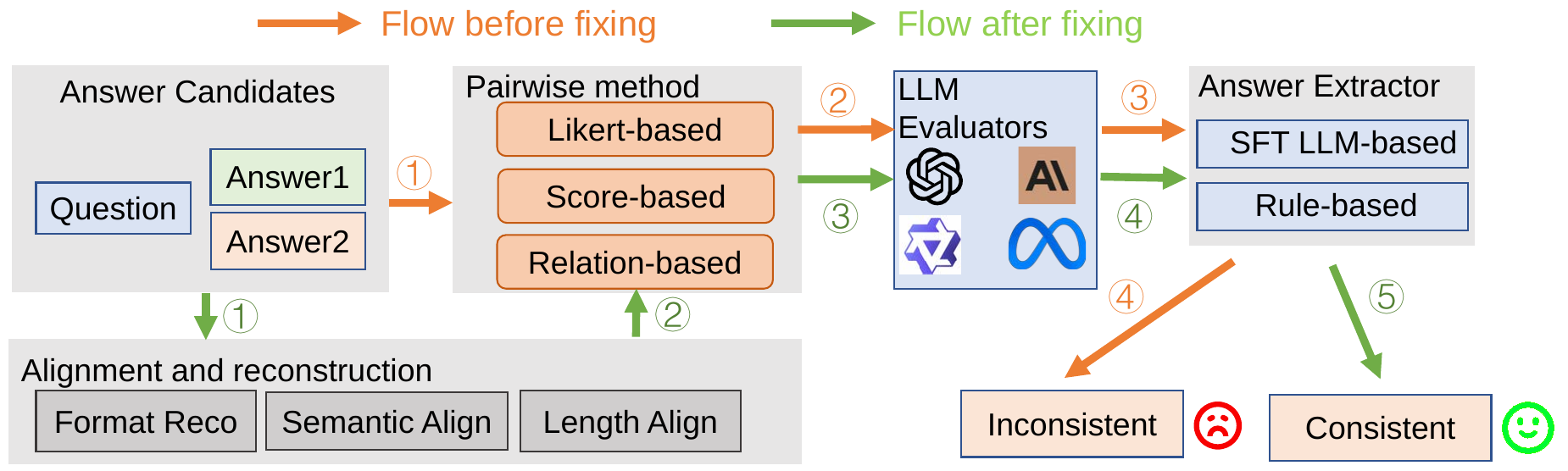}
    \caption{This is the overview of using \tool\ for LLM evaluation. ``Reco'' and ``SFT'' are short for ``recognition'' and ``supervised fine-tuning'', respectively.}
    \label{fig:overview}
\end{figure*}

This section explains the full pipeline of utilizing \tool\ for LLM evaluation.
As depicted in~\F~\ref{fig:overview}, typically there are four key steps: (1) \textbf{Data preparation}, (2) \textbf{Comparison method selection}, (3) \textbf{Evaluation}, and (4) \textbf{Answer extract}.

In the first step, we prepare the data for evaluation, which includes the questions and corresponding answers from two different LLMs to be compared.
If \tool\ is not implemented, we next choose the comparison method and formulate the input prompt, which has a great impact on the evaluation results, as we discussed in~\S~\ref{subsec:main-results}. The selected LLM evaluator is then queried with this prompt to obtain judgments.
Note that the raw evaluation results require additional processing for two reasons: (1) the output format may differ from the desired final verdicts and (2) the LLM evaluators may deviate from expected responses. For example, the LLM evaluator may fail to return the likert score for some questions but instead return the final verdict directly.
Therefore, we design an answer extractor to extract the final verdict from the evaluation results. Specifically, we adopt a hybrid method to extract the final verdict, which first tries to extract with a rule-based system, and if it fails, then it tries with a supervised fine-tuning Chatglm2~\citep{chatglm2} model.

The \tool-enhanced evaluation would necessitate an additional step of alignment and reconstruction, which constitutes the core of our framework. As elucidated in the main text, this procedure is vital for assessing the LLM answers' quality with less position bias.

\parh{Community needs.}
Notably, some leading LLM leaderboards, such as Chatbot Arena~\cite{zheng2023judging}, still rely on human votes for evaluation, which limits their scalability and increases costs. We believe that our method can be seamlessly integrated into these leaderboards to support an automatic evaluation process, significantly reducing costs and improving the scalability of the leaderboards. For scenarios where LLM-based evaluators are used, the community can easily implement \tool\ by altering the query construction in existing pipelines. By adding the splitting and merging steps, the community can use the consistent results from \tool\ as the final output. 

\section{Real-World Performance and Cost Analysis}
\label{app:real-world}

In this section, we provide a the performance and cost analysis of different LLM evaluators before and after fix by \tool\ in real-world settings. Notably, the carbon emission of GPT-3.5 is estimated following~\cite{chien2023reducing}.
We estimate the cost using the official pricing for cloud APIs and the Azure ND A100 v4 instances for local models.

\begin{table*}[!htpb]

    \footnotesize
    \centering
    
    \begin{tabular}{lrrrrr}
    \hline
    & \multirow{2}{*}{\begin{tabular}[r]{@{}r@{}}AR origin\\ (\%)\end{tabular}} & 
    \multirow{2}{*}{\begin{tabular}[r]{@{}r@{}}AR fix\\ (\%)\end{tabular}} &
    \multirow{2}{*}{\begin{tabular}[r]{@{}r@{}}Carbon Emitted \\ (CO$_2$eq / per 1k)\end{tabular}} &
    \multirow{2}{*}{\begin{tabular}[r]{@{}r@{}}Avg Cost \\ (USD / per 1k)\end{tabular}} & 
    \multirow{2}{*}{\begin{tabular}[r]{@{}r@{}}Avg Time \\ (s / per 1k)\end{tabular}} 
     \\ 
    &  &  & &\\\hline 
    GPT-4 & - & - & N/A & 29.78 & 13,446\\ \hline 
    GPT-3.5 & 82.50 & 88.59 & 7.22 & 2.85 & 2,192\\
    Qwen & 60.83 & 69.58 & N/A & 35.49 & 6,083\\
    Chatglm2 & 20.34 & 39.16 & 2.15 & 4.09 & 1,983\\
    Claude2 & 43.44 & 75.09 & N/A & 27.17 & 11,561\\
    \hline
    \end{tabular}
    \caption{Real-world comparison of different LLM evaluators' results before and after fix by \tool\ with that of GPT-4, including resource consumption. ``AR''
    denotes the agreement rate with GPT-4.
    }
    \label{tab:agreement-cost}

\end{table*}

\section{LLM Details}
\label{app:llm-details}

In this section, we provide more details about the LLM evaluators and answers used in our experiments.

\parh{LLM Evaluators.}~As introduced in~\S~\ref{subsec:settings}, we include both locally deployable
models that are open-source and proprietary models that are accessed through
only cloud APIs as LLM evaluators. 
For local models, we select Chatglm2~\citep{chatglm2} and Llama2~\citep{touvron2023llama2}, due to their
notable efficacy and convenient local deployment capabilities. For cloud-based
LLMs, we use GPT (including both GPT-4 and GPT-3.5)~\citep{openai2023gpt4} from
OpenAI, Qwen~\citep{qwen} from Alibaba, and Claude2~\citep{claude2} from
Anthropic. 
The rationale for using these models is based on their exceptional performance, since they are considered among the most advanced and powerful in the world. 
Specifically, we evaluate \tool\ using six distinct LLMs as evaluators:
\begin{itemize}
    \item \textbf{GPT-4}~\citep{openai2023gpt4} is a large multimodal model capable of processing image and text inputs to generate text outputs. GPT-4 demonstrates human-level aptitude on various professional and academic benchmarks. We utilize the 8K context length ``gpt-4-0613'' configuration by default. 
    \item \textbf{GPT-3.5}~ is a 175B parameter model from OpenAI offered in 4K and 16K context length versions. Our experiments use the 4K context ``gpt-3.5-turbo-0301'' model as default.
    \item \textbf{Claude2}~\citep{claude2} is the latest large language model released by Anthropic. It supports at most 100k tokens as input. We leverage the default Claude2 API in our tests.
    \item \textbf{Llama2}~\citep{touvron2023llama2}, an open-source series of LLMs from Meta AI ranging from 7B to 70B parameters, is trained on 2 trillion tokens and doubles Llama1's context length. Its fine-tuned iterations utilize over 1 million human annotations. We evaluate both 7B and 13B Llama2 chat models.
    \item \textbf{Qwen}~\citep{qwen} is a partially open-sourced LLM model released by Alibaba. We use the default API service provided by Alibaba cloud in our experiments. 
    \item \textbf{Chatglm2}~\citep{chatglm2} is the second-generation version of the open-source bilingual chat model ChatGLM-6B. We use the offered 6B version in our experiments.
\end{itemize}

\parh{LLM Answers.}~As mentioned in~\S~\ref{subsec:settings}, we consider eight answer combinations from different LLMs, specifically, the pairs are:
``gpt-3.5-turbo'' versus ``claude-v1'',  ``llama-13b'' versus ``vicuna-13b'', ``alpaca-13b'' versus ``vicuna-13b'', ``gpt-3.5-turbo'' versus ``gpt-4'', ``gpt-4'' versus ``claude-v1'', ``vicuna-13b'' versus ``vicuna-7b'', ``vicuna-7b'' versus ``alpaca-13b'', and ``gpt-4'' versus ``vicuna-13b''.
The answers are generated by the LLMs without any post-processing, and we reuse these answers from previous work~\citep{zheng2023judging}.
Notably, there is indeed some overlap between the models used as LLM evaluators and models used to generate the answers.

\section{Algorithm Illustration}
\label{app:split-algo}

\begin{figure*}[!htpb]
    \centering
    \includegraphics[width=1.0\textwidth]{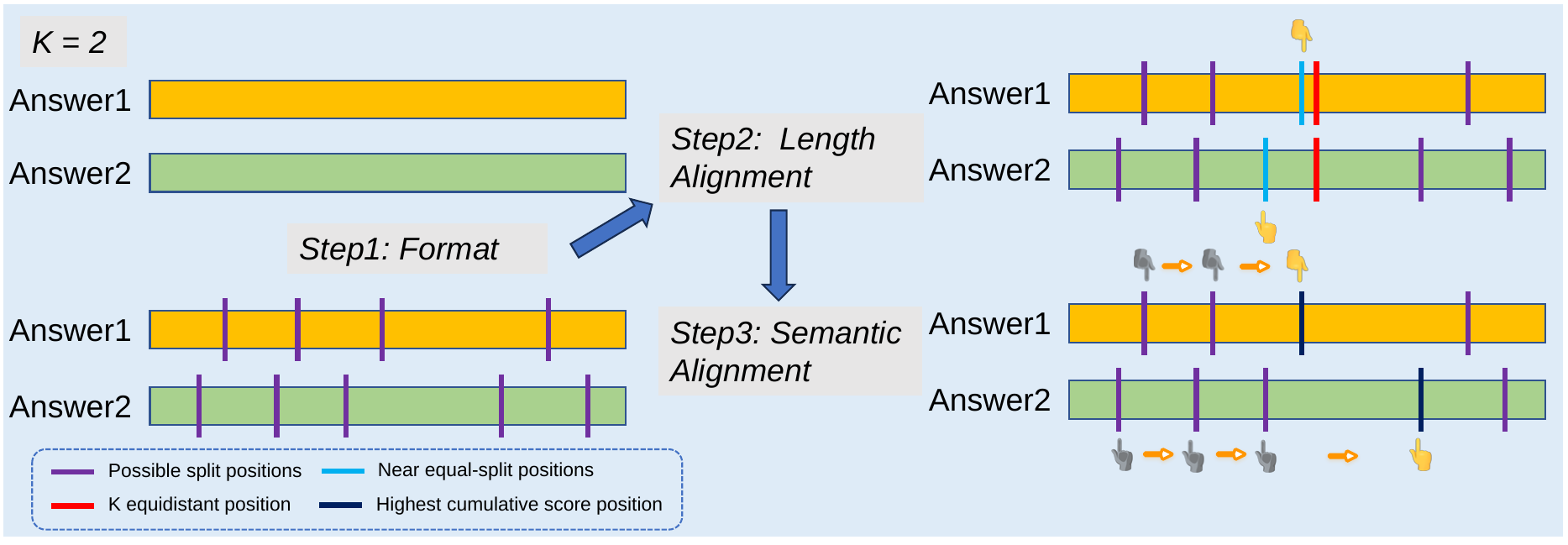}
    \caption{Schematic illustration of the proposed splitting algorithm, depicting its operation when configured with $k=2$ (i.e., division into two parts).  }
    \label{fig:algo-illustration}
\end{figure*}

To elucidate the operational details of the proposed splitting algorithm, we provide a schematic depiction in~\F~\ref{fig:algo-illustration}. 
Given two LLM-generated answers, the algorithm first identifies all candidate split positions coinciding with sentence boundaries in each answer. It then performs length alignment by initially dividing each answer equally into $k$ partitions and recording the corresponding split positions. Next, for each answer, the split position closest to the recorded locations is selected from the candidate positions. The answers are partitioned into $k$ segments at these chosen split positions. The resulting segments are fed as inputs to the LLM evaluator to obtain the respective judgments.

In cases where inconsistent judgments persist, the algorithm proceeds with semantic alignment to further divide each answer into $k$ parts. Specifically, an iterative search is conducted for optimal split positions that maximize the cumulative semantic similarity between the corresponding segments from the two answers. This traversal terminates when the complete set of potential split positions has been evaluated. Through this process based on both length and semantic alignment, the algorithm is able to decompose the LLM answers into aligned parts for more consistent and reliable evaluation.

\begin{algorithm*}[!htbp]
    \footnotesize
    \caption{Details of Step 3: Semantic Alignment ($k=2$)}\label{alg:partition}
    \tcc{Step3: semantic alignment.}
    $s_{max} = 0$, $r_1^{bestparts}=[], r_2^{bestparts}=[]$\\
        \For{$i$ in $range(len(r_1^{positions}))$}{
            \For{$j$ in $range(len(r_2^{positions}))$}{
                $pos_1 = r_1^{positions}[i], pos_2 = r_2^{positions}[j]$ \\
                $r_1^{parts}[0] = r_1[:pos_1], r_2^{parts}[0] = r_2[:pos_2]$ \\
                $r_1^{parts}[1] = r_1[pos_1:], r_2^{parts}[1] = r_2[pos_2:]$

                $s_{cum} = \sum_{i=1}^{2} similarity(r_1^{parts}[i], r_2^{parts}[i])$\\
                \tcc{Update max similarity score, keep best split positions.}
                \If{$s_{cum} > s_{max}$}{
                    $s_{max} = s_{cum}, r_1^{bestparts}=r_1^{parts}, r_2^{bestparts}=r_2^{parts}$
                    }
            }
        }
\end{algorithm*}

\begin{algorithm*}[!htbp]
    \footnotesize
    \caption{Details of Step 3: Semantic Alignment ($k=3$)}\label{alg:partition}
    \tcc{Step3: semantic alignment.}
    $s_{max} = 0$, $r_1^{bestparts}=[], r_2^{bestparts}=[]$\\
        \For{$i_1$ in $range(len(r_1^{positions}))$}{
            \For{$i_2$ in $range(i_1+1, len(r_1^{positions}))$}{

                \For{$j_1$ in $range(len(r_2^{positions}))$}{
                    \For{$j_2$ in $range(j_1+1, len(r_2^{positions}))$}{

                    $pos_{11} = r_1^{positions}[i_1], pos_{21} = r_2^{positions}[j_1]$ \\
                    $pos_{12} = r_1^{positions}[i_2], pos_{22} = r_2^{positions}[j_2]$ \\
                    $r_1^{parts}[0] = r_1[:pos_{11}], r_2^{parts}[0] = r_2[:pos_{21}]$ \\
                    $r_1^{parts}[1] = r_1[pos_{11}:pos_{12}], r_2^{parts}[1] = r_2[pos_{21}:pos_{22}]$ \\
                    $r_1^{parts}[2] = r_1[pos_{12}:], r_2^{parts}[2] = r_2[pos_{22}:]$ \\

                    $s_{cum} = \sum_{i=1}^{3} similarity(r_1^{parts}[i], r_2^{parts}[i])$\\
                    \tcc{Update max similarity score, keep best split positions.}
                    \If{$s_{cum} > s_{max}$}{
                        $s_{max} = s_{cum}, r_1^{bestparts}=r_1^{parts}, r_2^{bestparts}=r_2^{parts}$
                        }
                    }
                }
            }
        }
\end{algorithm*}

\section{LM Metric}
\label{app:lm-metric}

In this section, we first introduce the LM metric used in our experiments. 
Then we conduct a controlled experiment to find the optimal number of splits $k$ across different metrics in 
terms of performance and efficiency.

\parh{LM Metric.}~To clarify, we use the Sentence-BERT~\citep{reimers-2019-sentence-bert} to
measure the similarity between pairs. Sentence-BERT is a modification of the
pretrained BERT~\citep{bert} network that uses siamese and triplet network
structures to derive semantically meaningful sentence embeddings that can be
compared using cosine-similarity. We do not follow previous work~\citep{li2022cctest,li2023feasibility,DBLP:journals/tosem/GaoGHZNXL23} where CodeBLEU is used, as Sentence-BERT is efficient while maintaining the accuracy of BERT.

\parh{Efficiency Evaluation.}~We use the same setup as
in~\S~\ref{subsec:settings} to conduct the experiment. According to the
theoretical analysis in~\S~\ref{subsec:efficiency}, we set $k \in \{1, 2, 3,
4\}$ and evaluate their efficiency, the results are shown
in~\T~\ref{tab:lm-metric}. Note that $k$ is the number of segments after
splitting, thus $k=1$ means no splitting would be performed, which leads to 0 in
terms of execution time. In short, it can be interpreted from the table that the
execution time grows exponentially with the increasing $k$.

\begin{table}[!htpb]
    \footnotesize
    \setlength{\tabcolsep}{6pt}
    \centering
    \renewcommand{\arraystretch}{1}
    \begin{tabular}{lllll}
    \toprule[1.1pt]
                     & $k=1$     & $k=2$  & $k=3$     & $k=4$   \\ \midrule[.9pt]
    Token-overlap    &    0      & 0.31  & 3.71     & 33.12     \\
    Bert-model       &    0      & 2.37  & 21.3     & 295.10      \\ \bottomrule[1.1pt]
    \end{tabular}
    \caption{Average execution time per input of different metrics with different $k$.  }
    \label{tab:lm-metric}
\end{table}

\begin{table}[!htpb]
    \footnotesize
    \setlength{\tabcolsep}{6pt}
    \centering
    \renewcommand{\arraystretch}{1}
    \begin{tabular}{lllll}
    \toprule[1.1pt]
                     & $k=1$     & $k=2$  & $k=3$   &   $k=4$   \\ \midrule[.9pt]
    Token-overlap    &    -      & 53.3  & 66.7     &  73.3 \\
    Bert-model       &    -      & 55.9  & 66.7     &  66.7   \\ \bottomrule[1.1pt]
    \end{tabular}
    \caption{Fixed coverage rates of different metrics with different $k$.  }
    \label{tab:lm-perf}
\end{table}

\parh{Performance Evaluation.}~Following the experimental setup described above, we set $k \in \{1, 2, 3, 4\}$ and evaluate their performance. 
To clarify, we use the answers from the LLM ``gpt-3.5-turbo'' and ``claude-v1'' in our experiments (under the same conditions outlined in~\S~\ref{subsec:human-study}), where in total of 80 questions are fed to GPT-3.5 for evaluation.
The results are shown in~\T~\ref{tab:lm-perf}, where we can see that with the
increasing $k$, the fixed coverage rates of both metrics are increasing, and when $k=3$, the
fixed coverage rate of both metrics is the same, which is 66.7\%. 
However, further increasing $k$ results in limited additional gains in coverage.
Considering the execution time which grows exponentially with the increasing $k$, we choose
$k=3$ with token-overlap as the default setting in our experiments.
Finally, we conclude that the semantic similarity score used in \tool\ achieves a good balance between coverage and execution time.

\section{On Llama2}
\label{app:bad-example}

Although LLMs can be effective evaluators, they are not infallible. In some cases, LLM evaluators may struggle to provide a meaningful verdict.
We find that these models would become too conservative to make a verdict, no matter how the answers are split and
aligned. 
Below is an example of the Llama2 models~\citep{touvron2023llama2} that are not well-served as evaluators.

\begin{table}[!htbp]
    \small
    \begin{tcolorbox}

    Assistant A: Alpaca-13b, Assistant B: Vicuna-13b

    [Question] \textcolor[rgb]{0,0,0.9}{\{How can I improve my time management skills?\}}
    
    \textcolor[rgb]{0.2,0.4,0.2}{[The Start of Assistant A's response]} \textcolor[rgb]{0,0,0.9}{\\Improving time management skills involves setting priorities, breaking tasks into smaller chunks, delegating tasks, avoiding multitasking, and taking regular breaks. Additionally, it is important to have a positive attitude and be realistic with goals. Making a to-do list, tracking time, and using technology to automate mundane tasks can also help improve time management skills.\\} \textcolor[rgb]{0.2,0.4,0.2}{[The End of Assistant A's response]}
    
    \textcolor[rgb]{0.2,0.4,0.2}{[The Start of Assistant B's response]} \textcolor[rgb]{0,0,0.9}{\\Improving your time management skills can help you to be more productive, focused, and less stressed. Here are some tips to help you improve your time management skills: 1. Set clear goals: Identify what you want to achieve and set specific, measurable, and realistic goals. Prioritize your tasks and focus on the most important ones first. 2. Use a calendar or planner: Use a tool to organize your time and schedule your tasks. This can help you to stay on track and make the most of your time. 3. Break down tasks: Break down larger tasks into smaller, more manageable steps. This can help you to focus on one step at a time and make steady progress. 4. Eliminate distractions: Identify and eliminate distractions that may be preventing you from being productive. This can include turning off your phone, closing unnecessary tabs on your computer, or finding a quiet place to work. 5. Practice self-discipline: Use self-discipline to stay focused and avoid procrastination. Set a timer for each task and challenge yourself to complete it before the timer goes off. 6. Take breaks: Take regular breaks to rest and recharge. This can help you to stay focused and avoid burnout. 7. Review your progress: Regularly review your progress and adjust your strategy as needed. This can help you to stay on track and make improvements over time.\\
    } \textcolor[rgb]{0.2,0.4,0.2}{[The End of Assistant B's response] }
    \\ \\
    Evaluator: \textcolor[rgb]{0.9,0.3,0.5}{\{Llama2-7b-chat\}},
    Form: \textcolor[rgb]{0.9,0.3,0.5}{\{score-based\}},
    Assessment: \textcolor[rgb]{0.9,0.3,0.5}{\{\textbackslash{}n\textbackslash{}n\textbackslash{}n\textbackslash{}n\textbackslash{}n\textbackslash{}n\textbackslash{}n\textbackslash{}n\textbackslash{}n  \}} \\
    Evaluator: \textcolor[rgb]{0.9,0.3,0.5}{\{Llama2-7b-chat\}},
    Form: \textcolor[rgb]{0.9,0.3,0.5}{\{likert-based\}},
    Assessment: \textcolor[rgb]{0.9,0.3,0.5}{\{""\}} \\
    Evaluator: \textcolor[rgb]{0.9,0.3,0.5}{\{Llama2-13b-chat\}},
    Form: \textcolor[rgb]{0.9,0.3,0.5}{\{score-based\}},
    Assessment: \textcolor[rgb]{0.9,0.3,0.5}{\{Please proceed with your evaluation.\}} \\
    Evaluator: \textcolor[rgb]{0.9,0.3,0.5}{\{Llama2-13b-chat\}},
    Form: \textcolor[rgb]{0.9,0.3,0.5}{\{likert-based\}},
    Assessment: \textcolor[rgb]{0.9,0.3,0.5}{\{\textbackslash{}n\textbackslash{}nPlease provide your feedback.\}} \\

\end{tcolorbox}
\caption{The assessments of Llama2-7b-chat and Llama2-13b-chat on two comparison forms.}
\label{tab:llama2-example}
\end{table}

\section{Prompt Templates}
\label{app:prompt-template}

\subsection{Comparison Forms}
\label{app:comparison-template}

In this section, we provide the detailed templates for the three comparison forms, including relation-based (\T~\ref{tab:rela-template}), score-based (\T~\ref{tab:score-template}), and likert-based forms (\T~\ref{tab:likert-template}).

\begin{table}[!htbp]
    \small
    \begin{tcolorbox}
    
    [Question] \textcolor[rgb]{0,0,0.9}{\{Q\}}
    
    [The Start of Assistant A's response] \textcolor[rgb]{0,0,0.9}{\{R1\}} [The End of Assistant A's response]
    
    [The Start of Assistant B's response] \textcolor[rgb]{0,0,0.9}{\{R2\}} [The End of Assistant B's response] 
    
    [System] 
    
    Please act as an impartial judge and evaluate the quality of the responses provided by two AI assistants to the user question displayed below. 
    
    You should choose the assistant that follows the user's instructions and answers the user's question better. Your evaluation should consider factors such as the helpfulness, relevance, accuracy, depth, creativity, and level of detail of their responses. Begin your evaluation by comparing the two responses and provide a short explanation. 
    
    Avoid any positional biases and ensure that the order in which the responses were presented does not influence your decision. Do not allow the length of the responses to influence your evaluation. Do not favor certain names of the assistants. Be as objective as possible. 
    
    After providing your explanation, output your final verdict by strictly following this format: [[A]] if assistant A is better, [[B]] if assistant B is better, and [[C]] for a tie.

\end{tcolorbox}
\caption{The relation-based evaluation form with three slots (\textcolor[rgb]{0,0,0.9}{\{Q\}, \{R1\} and \{R2\}}) from \citep{zheng2023judging}.}
\label{tab:rela-template}
\end{table}

\begin{table}[!htbp]
    \small
    \begin{tcolorbox}
    
    [Question] \textcolor[rgb]{0,0,0.9}{\{Q\}}
    
    [The Start of Assistant A's response] \textcolor[rgb]{0,0,0.9}{\{R1\}} [The End of Assistant A's response]
    
    [The Start of Assistant B's response] \textcolor[rgb]{0,0,0.9}{\{R2\}} [The End of Assistant B's response] 
    
    [System] 
    
    We would like to request your feedback on the performance of two AI assistants in response to the user question displayed above.
    
    Please rate the helpfulness, relevance, accuracy, level of details of their responses. Each assistant receives an overall score on a scale of 1 to 10, where a higher score indicates better overall performance.
    
    Please first output a single line containing only two values indicating the scores for Assistant A and B, respectively. The two scores are separated by a space. In the subsequent line, please provide a comprehensive explanation of your evaluation, avoiding any potential bias and ensuring that the order in which the responses were presented does not affect your judgment.

\end{tcolorbox}
\caption{The score-based evaluation form with three slots (\textcolor[rgb]{0,0,0.9}{\{Q\}, \{R1\} and \{R2\}}).}
\label{tab:score-template}
\end{table}

\begin{table}[!htbp]
    \small
    \begin{tcolorbox}
    
    [Question] \textcolor[rgb]{0,0,0.9}{\{Q\}}
    
    [The Start of Assistant A's response] \textcolor[rgb]{0,0,0.9}{\{R1\}} [The End of Assistant A's response]
    
    [The Start of Assistant B's response] \textcolor[rgb]{0,0,0.9}{\{R2\}} [The End of Assistant B's response] 
    
    [System] 
    
    We would like to request your feedback on the performance of two AI assistants in response to the user question displayed above.
    
    Please compare the helpfulness, relevance, accuracy, level of details of their responses. 

    The rating should be from the set of 1, 2, 3, 4, 5, 6, or 7, where higher numbers indicated that Assistant A was better than Assistant B. Please first output a single line containing only one value indicating the preference between Assistant A and B. 
    
    In the subsequent line, please provide a brief explanation of your evaluation, avoiding any potential bias and ensuring that the order in which the responses were presented does not affect your judgment.

\end{tcolorbox}
\caption{The likert-based evaluation form with three slots (\textcolor[rgb]{0,0,0.9}{\{Q\}, \{R1\} and \{R2\}}).}
\label{tab:likert-template}
\end{table}

\subsection{Alignment Templates}
\label{app:alignment-template}

In this section, we provide a detailed example in~\T~\ref{tab:tool-template}, demonstrating how to align the responses from two assistants with three slots.

\begin{table*}[hbp]
    \small
    \begin{tcolorbox}
    
    [Question] \textcolor[rgb]{0,0,0.9}{\{What are the most effective ways to deal with stress?\}}
    
    \textcolor[rgb]{0.2,0.4,0.2}{[The Start of Assistant A's response part 1]} \textcolor[rgb]{0,0,0.9}{\\Here are some effective ways to deal with stress:\\
    1. Exercise regularly: Physical activity can help reduce stress and improve mood.\\} \textcolor[rgb]{0.2,0.4,0.2}{[The End of Assistant A's response part 1]}
    
    \textcolor[rgb]{0.2,0.4,0.2}{[The Start of Assistant B's response part 1]} \textcolor[rgb]{0,0,0.9}{\\Here are some of the most effective ways to deal with stress:\\
    • Exercise regularly. Exercise is a great way to release pent up stress and tension. Even taking a 30 minute walk can help clear your mind and relieve stress.\\
    } \textcolor[rgb]{0.2,0.4,0.2}{[The End of Assistant B's response part 1] }

    \textcolor[rgb]{0.2,0.4,0.2}{[The Start of Assistant A's response part 2]} \textcolor[rgb]{0,0,0.9}{\\2. Practice relaxation techniques: Deep breathing, meditation, and yoga are some effective ways to relax and reduce stress levels.\\
    3. Get enough sleep: Lack of sleep can increase stress levels. Try to get at least 7-8 hours of sleep each night.\\
    4. Eat a healthy diet: A well-balanced diet can help reduce stress levels and improve overall health.\\} \textcolor[rgb]{0.2,0.4,0.2}{[The End of Assistant A's response part 2]}

    \textcolor[rgb]{0.2,0.4,0.2}{[The Start of Assistant B's response part 2]} \textcolor[rgb]{0,0,0.9}{\\• Practice deep breathing and relaxation techniques. Things like meditation, yoga, and mindfulness exercises can help lower your stress levels. Taking some deep breaths can help you calm down in the moment.\\
    • Get enough sleep. Aim for 7 to 8 hours of sleep per night. Lack of sleep can intensify stress. Sleep helps your body and mind rest and recharge.\\
    • Eat a healthy, balanced diet. A healthy diet can boost your mood and reduce stress. Focus on whole foods like fruits, vegetables, whole grains, and lean protein. Limit caffeine and alcohol intake.\\
    • Spend time with others. Social support from family and friends can help reduce stress levels. Connecting to others can help take your mind off your worries and boost your mood.\\} \textcolor[rgb]{0.2,0.4,0.2}{[The End of Assistant B's response part 2]}

    \textcolor[rgb]{0.2,0.4,0.2}{[The Start of Assistant A's response part 3]} \textcolor[rgb]{0,0,0.9}{\\5. Connect with others: Spending time with friends and family, joining a support group, or participating in social activities can help reduce stress levels and improve mood.\\
    6. Time management: Prioritizing tasks and creating a schedule can help reduce stress and increase productivity.\\
    7. Seek professional help: If stress is impacting your daily life, consider seeking professional help such as therapy or counseling.\\
    } \textcolor[rgb]{0.2,0.4,0.2}{[The End of Assistant A's response part 3]}

    \textcolor[rgb]{0.2,0.4,0.2}{[The Start of Assistant B's response part 3]} \textcolor[rgb]{0,0,0.9}{\\• Limit unhealthy habits. Things like smoking, excessive drinking, or drug use can worsen symptoms of stress. Try to avoid these unhealthy habits as much as possible.\\
    • Practice mindfulness. Spending time each day being fully present and aware of your thoughts and feelings can help reduce stress and increase feelings of calm. Try yoga, meditation, or simply sit in a quiet space for a few minutes.\\
    • Seek professional help if needed. If you are dealing with chronic or severe stress, it may help to talk to a therapist. A counselor can provide advice and coping strategies tailored to your situation. Therapy and counseling can be very effective at reducing stress and anxiety.\\
    • Take periodic breaks. It is important to take time each day to unwind and do something you enjoy. Read a book, take up a hobby, or simply relax in a bath. Taking breaks prevents burnout and recharges your body and mind.\\} \textcolor[rgb]{0.2,0.4,0.2}{[The End of Assistant B's response part 3]}

    [System] 
    
    Please act as an impartial judge and evaluate the quality of the responses provided by two AI assistants to the user question displayed below. 
    
    You should choose the assistant that follows the user's instructions and answers the user's question better. Your evaluation should consider factors such as the helpfulness, relevance, accuracy, depth, creativity, and level of detail of their responses. Begin your evaluation by comparing the two responses and provide a short explanation. 
    
    Avoid any positional biases and ensure that the order in which the responses were presented does not influence your decision. Do not allow the length of the responses to influence your evaluation. Do not favor certain names of the assistants. Be as objective as possible. 
    
    After providing your explanation, output your final verdict by strictly following this format: [[A]] if assistant A is better, [[B]] if assistant B is better, and [[C]] for a tie.

\end{tcolorbox}
\caption{The detailed prompt illustrated in~\F~\ref{fig:example}. We use
relation-based form to construct the system prompt. The prompt in
\textcolor[rgb]{0.2,0.4,0.2}{green} is the ``split boundary prompts''.}
\label{tab:tool-template}
\end{table*}

\section{Generalizability of \name}
\label{app:generalizability}

\subsection{Extended Open-Ended Questions}
\label{subsec:extended-open-ended-questions}

To evaluate the generalizability of \name, we first generate an extended set of open-ended questions based on the original MT-Bench dataset. As introduced in \S~\ref{subsec:settings}, MT-Bench contains 80 elaborated open-ended questions spanning different categories. Following their approach, we use each question in the original dataset as a seed and ask GPT-4 to generate a number of similar questions according to its category.
The specific prompt used is: ``You are given a problem whose category is \{category\}, please generate \{number\_example\} problems.'', where ``number\_example'' is set to 10. It is worth noting that the generated questions may have duplicates. Therefore, whenever a new question is generated, we check whether it is a duplicate of any existing questions. If so, we discard it directly. 

\begin{figure}[!htpb]
    \centering
    \begin{subfigure}[b]{0.48\textwidth}
        \centering
        \includegraphics[width=\textwidth]{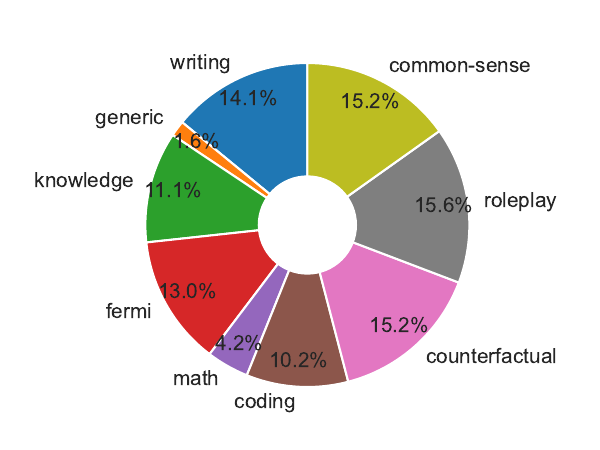}
        \caption{Distribution of Question Categories}
        \label{fig:category-distribution}
    \end{subfigure}
    \hfill
    \begin{subfigure}[b]{0.48\textwidth}
        \centering
        \includegraphics[width=\textwidth]{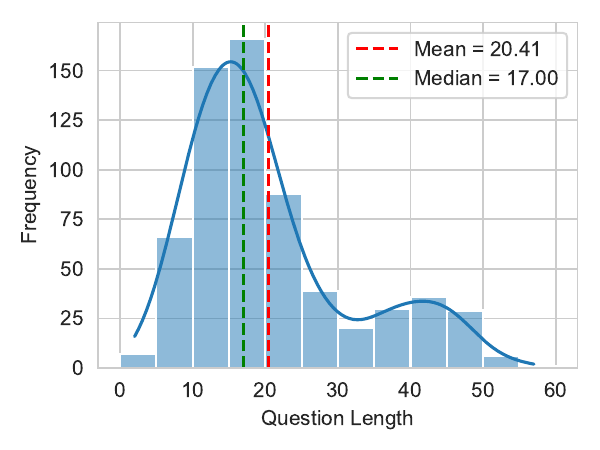}
        \caption{Distribution of Number of Question Words}
        \label{fig:word-distribution}
    \end{subfigure}
    \caption{Analysis of the extended open-ended question dataset: (a) distribution of question categories and (b) distribution of the number of question words.}
    \label{fig:dataset-analysis}
\end{figure}

\begin{table}[!htbp]
    \centering
    \begin{tabular}{l|r|r|r}
    \hline
    & GPT-4 & GPT-3.5 & Avg Cost \\ \hline
    VANILLA & 52.7 & 44.4 & 1x \\ 
    Our baseline & 60.0 & 55.0 & 1.03x \\ 
    MEC & 60.9 & 55.6 & 3.29x \\ 
    MEC+BPC  & 62.5 & 58.7 & 3.29x \\ \hline
    Ours & 65.0 & 63.8 & 1.68x \\ \hline
    HITL & 73.8 & 71.3 & 97.3x \\ \hline
    \end{tabular}
    \caption{Comparison of different methods and their performance on GPT-4 and GPT-3.5, along with the average cost.}
    \label{tab:comparison-strongerbaseline}
\end{table}

Using the above method, we obtain a total of 640 questions as our new dataset. We further evaluate this extended dataset from two aspects. First, we analyze the distribution of question categories, as shown in \F~\ref{fig:category-distribution}. We can observe that our generated dataset has coverage across all categories. Secondly, we analyze the distribution of the number of question words, as shown in \F~\ref{fig:word-distribution}. The vast majority of questions have lengths between 10 and 20 words, with mean and median values of 20.41 and 17.00, respectively.

\subsection{Main Results}
\label{subsec:main-results-generalizability}

After obtaining the extended dataset, we further conducted experiments on it to validate the effectiveness of \tool. The experimental settings are similar to those introduced in \S~\ref{subsec:settings}. In the first step, we queried three models: GPT-3.5-turbo, DeepSeek-v2~\cite{deepseekai2024deepseekv2}, and Meta-LLaMA-3-70b-instruct~\cite{llama3}. The first model has been introduced in the previous experiments, while the second and third are relatively new open-source models with capabilities similar to GPT-3.5-turbo, especially in code-related tasks~\citep{wang2024exploring}. We selected GPT-3.5 and GPT-4o as our evaluators to assess the quality of the generated questions. We experimented with three pairwise comparison methods: score-based, Likert-based, and relation-based.

Our experimental results are presented in \T~\ref{tab:extented-result}. We can observe that \tool\ is able to improve consistency under all evaluators and comparison methods. Most of the findings are consistent with those in the main text. For example, more advanced models have higher average consistency rates. At the same time, we observe that GPT-4o, despite being a newer model compared to GPT-4, still has a lower consistency rate in the Likert-based comparison.

This indirectly demonstrates the necessity of our method, as we can see that with the updating of models, their biases do not decrease accordingly. In other words, a better model does not necessarily imply smaller biases. We also observe that as the performance of open-source models improves, the quality of their generated answers also increases to a level similar to that of GPT-3.5, making it more difficult for evaluators to assess. This highlights the necessity of our method from another perspective. Facing an increasing number of models, relying solely on human evaluators to assess answer quality is insufficient, while \tool\ can help us better automate the evaluation of answer quality.

\begin{table*}[!htbp]
    \small
    \footnotesize
    \centering
    \begin{tabular}{c | c | c | c | c | c}
        \toprule
        \textbf{Evaluators}& \textbf{De. Method}  & \textbf{Model} & \textbf{Relation-based} & \textbf{Score-based} & \textbf{Likert-based} \\
        \midrule

        \multirow{3}*{GPT-3.5} 
        & &\% Origin Con & 87.1 & 57.0 & 77.5   \\
        & API&\% \tool\ Con
        & \textbf{92.5(+6.19\%)}   
        & \textbf{63.3(+11.1\%)} 
        & \textbf{92.5(+19.4\%)}\\
        & &\% Fixed Coverage & 69.0 & 39.5 &  89.5  \\

        \midrule
        \multirow{3}*{GPT-4o} 
        & &\% Origin Con & 93.5 & 91.5 & 34.5   \\
        & API &\% \tool\ Con
        & \textbf{98.8(+5.7\%)} 
        & \textbf{99.1(+8.3\%)} 
        & \textbf{37.9(+9.8\%)} \\
        & &\% Fixed Coverage & 80 & 92.5 & 27.4  \\

        \bottomrule
    \end{tabular}
    \caption{The main results of \tool\ on the extended dataset. }
    \label{tab:extented-result}
\end{table*}

\begin{table*}[!htbp]
    \centering
    \resizebox{1.0\linewidth}{!}{
    \begin{tabular}{lp{8cm}p{8cm}}
    \hline
    \multicolumn{3}{c}{\textbf{Criteria for Assessing the Answer}} \\
    \hline
    \textbf{Dimension} & \textbf{Description} & \textbf{Main Checklist} \\
    \hline
    Feasibility & Answers should be clear, specific, feasible, and easily understandable. & Check for clarity, specificity, practicality, and comprehensibility. \\
    \hline
    Relevance & Answers should directly address the user's request, provide helpful solutions, faithfully answer the question, and are not off-topic. & Check for direct relevance to the question, helpfulness of solutions, faithfulness to the request, and staying on-topic. \\
    \hline
    Safety & Responses should be harmless and not about non-ethical topics. & Check for the absence of harmful or unethical content. \\
    \hline
    Correctness & Answers should not contain factual errors. & Check for the absence of factual inaccuracies or misinformation. \\
    \hline
    \end{tabular}
    }
    \caption{Human evaluation criteria for the better response.}
    \label{tab:human-eval-criteria}
    \end{table*}
\section{Error Analysis}
\label{app:error-analysis}

We do recognize that our method may not be able to perfectly resolve all inconsistencies, and there might be instances where it fails. In analyzing cases where \tool\ fails, we have found that it might not perform effectively when dealing with ethical issues. For example, when using the GPT-3.5 evaluator to assess responses to a question such as "Explain the process of gene editing using CRISPR-Cas9 technology, and discuss its potential applications and ethical implications," \tool\ may not be able to provide consistent answers (details are provided in~\T~\ref{tab:error-example}.) Through this analysis, we believe that our framework may have limitations on specific topics, especially when addressing moral issues related to human beings.

From a higher-level perspective on the design of \tool, it is important to note that the system does not engage in multiple rounds of querying to arrive at a consistent answer. Consequently, there will inevitably be instances where \tool\ is unable to reach a consistent final verdict, as exemplified in \T~\ref{tab:error-example}. In such cases, we advise users to exercise caution when considering the final verdict and, if necessary, seek the opinion of human experts for further evaluation, particularly when dealing with sensitive or complex subject matter.

\parh{Position Preference Analysis.}
We further analyze the position preferences of different LLM evaluators, both in their original judgments and after \tool\ enhancement. \T~\ref{tab:position-analysis} shows the percentage of responses in the first position selected as the final answer. We observe that different LLMs exhibit varying position biases, with GPT-4 favoring the first position (62.31\%) and GPT-3.5 preferring the second (66.44\%), which is also in line with previous work~\citep{wang2023large}. \tool-enhanced results generally show a more balanced selection, with percentages closer to 50\% for most LLMs. For instance, Claude2's first-position preference increases from 19.46\% to 32.04\%, while GPT-4's slightly decreases from 62.31\% to 60.85\%. These findings suggest that \tool\ effectively mitigates position bias while preserving some of the LLMs' intrinsic characteristics, thereby improving the consistency and fairness of the evaluation process.

\begin{table}[htbp]
    \centering
    \resizebox{0.8\linewidth}{!}{
    \begin{tabular}{l|c|c}
    \hline
    LLM Evaluator & Original & \tool \\
    \hline
    GPT-3.5 & 33.56\% & 39.12\% \\
    Chatglm2 & 48.38\% & 57.89\% \\
    Qwen & 36.73\% & 39.13\% \\
    Claude2 & 19.46\% & 32.04\% \\
    GPT-4 & 62.31\% & 60.85\% \\
    \hline
    \end{tabular}
    }
    \caption{Percentage of first-position responses selected as the final answer}
    \label{tab:position-analysis}
\end{table}

\begin{table*}[!htbp]
    \small
    \begin{tcolorbox}

    Assistant A: gpt-3.5-turbo, Assistant B: claude-v1
    
    \textcolor[rgb]{0.2,0.4,0.2}{[Question]} \textcolor[rgb]{0,0,0.9}{\{Explain the process of gene editing using CRISPR-Cas9 technology, and discuss its potential applications and ethical implications.\}}
    
    \textcolor[rgb]{0.2,0.4,0.2}{[Assistant A's response]} \textcolor[rgb]{0,0,0.9}{\\Gene editing is a process of changing the DNA of an organism in order to alter its traits. CRISPR-Cas9 is a gene editing technology that has revolutionized the field of genetic engineering. In this technique, the Cas9 enzyme is guided by a molecule called guide RNA (gRNA), which is programmed to recognize and cut specific sequences of DNA. This cut then triggers the cell\u2019s natural DNA repair mechanism, leading to the addition, deletion, or replacement of specific DNA sequences.\textbackslash{}n\textbackslash{}nCRISPR-Cas9 has a wide range of potential applications, including the ability to cure genetic diseases, develop new treatments for cancer, and create new crop varieties with improved yield and resistance to pests and disease. It can also be used to enhance human traits, such as intelligence or physical performance, which raises ethical concerns about the potential for eugenics.\textbackslash{}n\textbackslash{}nOne of the main ethical concerns associated with CRISPR-Cas9 is the risk of unintended consequences. Gene editing can have unforeseeable impacts on the environment and future generations, which could lead to potentially harmful and irreversible effects. There are also concerns about the potential misuse of gene editing technology for unethical purposes, such as creating designer babies or enhancing traits for non-therapeutic purposes.\textbackslash{}n\textbackslash{}nAnother ethical consideration is the issue of access and equity. CRISPR-Cas9 is an expensive technology, and its use could widen the gap between the haves and the have-nots, with potentially devastating consequences for global health and social justice.\textbackslash{}n\textbackslash{}nIn conclusion, while the potential benefits of CRISPR-Cas9 are immense, its ethical implications are complex and multifaceted. Careful consideration and regulation of its use will be required to maximize its benefits, while minimizing its potential risks and harms.\\} 
    
    \textcolor[rgb]{0.2,0.4,0.2}{[Assistant B's response]} \textcolor[rgb]{0,0,0.9}{\\CRISPR-Cas9 is a gene editing technology that allows scientists to precisely edit DNA sequences. It involves using a protein called Cas9 that can cut strands of DNA, coupled with a strand of RNA that can guide the Cas9 protein to a specific sequence of DNA. The steps in the CRISPR-Cas9 process are:\textbackslash{}n\textbackslash{}n1. Identify the target DNA sequence you want to edit. This is done by designing a strand of RNA that is complementary to that target sequence.\textbackslash{}n\textbackslash{}n2. Introduce the RNA strand and the Cas9 protein into the cell. The RNA will guide the Cas9 protein to the target DNA sequence.\textbackslash{}n\textbackslash{}n3. The Cas9 protein cuts both strands of the DNA at the target site.\textbackslash{}n\textbackslash{}n4. The cell's DNA repair mechanisms kick in to repair the cut. Scientists can introduce a new section of DNA to insert at the cut site, thus editing the DNA sequence. \textbackslash{}n\textbackslash{}n5. The edited DNA sequence becomes incorporated into the genome. In subsequent cell divisions, the edited sequence is passed on to daughter cells.\textbackslash{}n\textbackslash{}nCRISPR-Cas9 has many promising applications, including treating genetic diseases, improving crop yields, and bringing back extinct species. However, it also raises ethical issues, including:\textbackslash{}n\textbackslash{}n\u2022 Safety concerns about off-target effects. The CRISPR system could edit DNA at unwanted locations. \textbackslash{}n\textbackslash{}n\u2022 Germline editing concerns. Changes to DNA in reproductive cells could be passed on to future generations, raising ethical questions.\textbackslash{}n\textbackslash{}n\u2022 Enhancement concerns. CRISPR could potentially be used for human enhancements like increasing intelligence or muscle mass. This raises ethical questions about "designer babies."\textbackslash{}n\textbackslash{}n\u2022 Equality concerns. Gene editing may only be available to the wealthy, exacerbating inequality. \textbackslash{}n\textbackslash{}n\u2022 Slippery slope concerns. Some argue that even if germline editing is allowed in some circumstances, it could lead down a slippery slope towards more ethically questionable uses.\textbackslash{}n\textbackslash{}nThat covers the basic process of how CRISPR-Cas9 gene editing works and some of the main applications and ethical implications of this technology. Please let me know if you would like me to explain anything in more detail.\\
    } 
    \textcolor[rgb]{0.2,0.4,0.2}{[Judge answer in original order]} \textcolor[rgb]{0,0,0.9}{Both Assistant A and Assistant B provided detailed and accurate explanations of the CRISPR-Cas9 gene editing process and its potential applications and ethical implications. However, Assistant A provided a more comprehensive and nuanced discussion of the ethical implications, including concerns about unintended consequences, access and equity, and the potential for eugenics. Assistant A also provided a more complete summary of the potential applications of CRISPR-Cas9, including its use in cancer treatments and crop improvement. Therefore, I believe that Assistant A provided a better response to the user's question. \textbackslash{}n\textbackslash{}n[[A]].} \\
    \textcolor[rgb]{0.2,0.4,0.2}{[Judge answer after exchanging the order]} \textcolor[rgb]{0,0,0.9}{Both Assistant A and Assistant B provided detailed and informative responses to the user's question about CRISPR-Cas9 gene editing technology, its process, potential applications, and ethical implications. However, Assistant A provided a more complete and structured answer that directly addressed all aspects of the user's question, including the steps involved in the CRISPR-Cas9 process, its potential applications in treating genetic diseases, improving crop yields, and bringing back extinct species, as well as the ethical concerns associated with gene editing, such as safety concerns, germline editing, enhancement concerns, equality concerns, and slippery slope concerns. Assistant A also offered to provide more information if needed. On the other hand, Assistant B's response was also informative and covered some of the same topics, but it was less structured and did not directly address all aspects of the user's question. Therefore, based on the above evaluation, I would choose [[A]] as the better assistant. }
    
\end{tcolorbox}
\caption{Ethical issues where \tool\ fails to provide consistent answers, using GPT-3.5 evaluator.}
\label{tab:error-example}
\end{table*}

\section{Annotation Process}
\label{app:annotation-process}

In \S~\ref{subsec:human-study}, we engage five experts to assess the effectiveness of \tool-enhanced evaluations across 80 questions. Each participant receives an online questionnaire containing questions and their corresponding answers, without specifying the source of the answers. Prior to the questionnaire, participants are provided with concise instructions on the task and evaluation criteria. The relation-based comparison forms, as shown in \T~\ref{tab:rela-template}, are employed to compare the candidate answers. They are asked to evaluate the answers based on the criteria outlined in \T~\ref{tab:human-eval-criteria}. To maintain the quality of the evaluation and prevent fatigue, participants are required to take a break after assessing 40 questions, and a 30-second interval is set between each question.

To ensure a comprehensive evaluation, our relation-based comparison form includes an option for annotators to indicate when they consider two outputs to be of equal quality. Specifically, option [[C]] in \T~\ref{tab:rela-template} allows annotators to express that ``these two outputs are tied''. However, our analysis reveals that only 11.25\% of the responses are ultimately classified as ties, which aligns with findings from previous study~\citep{wang2023large}.

We also investigate the potential correlation between position inconsistency and human judgments of ties. A statistical analysis yields a p-value of 0.7280, indicating no significant correlation. This finding suggests that simply treating all instances of LLM position inconsistency as ties would not accurately reflect human judgments. 
It's important to note that treating all instances of LLM position inconsistency as ties would lead to an unrealistically high proportion of tied outcomes. In our analysis with Llama2, for instance, this approach would result in 63.59\% of the comparisons being classified as ties.

These results underscore the importance of a more specific approach to resolving LLM position inconsistencies rather than broadly categorizing them as ties. Our approach aims to provide a more accurate reflection of quality differences between outputs, even in cases where LLMs initially show inconsistency in their evaluations.

\section{Stronger Baselines}
\label{app:stronger-baseline}

To further demonstrate the effectiveness of \name, we compare our framework with various baselines of the traditional LLM evaluator setup, including the vanilla method, MEC, BPC, and HITLC. Specifically, the VANILLA method simply asks evaluators to output their preferences without any explanation. MEC, BPC, and Human-in-the-loop (HITL) are methods proposed in~\cite{wang2023large}, requiring multiple turns of querying or human effort.

We present the results in~\T~\ref{tab:comparison-strongerbaseline}, showing the agreement rate between humans and corresponding LLM evaluators for GPT-4 and GPT-3.5.

From the table, we can observe that our framework outperforms all methods except for HITLC, which requires human effort at an extremely high cost. Given that our framework is fully automated and low-cost, we believe that it serves as a strong baseline for future research.

\end{document}